\documentclass{article} \usepackage{iclr2027_conference,times}

\usepackage{amsmath,amsfonts,bm}

\def\eqref#1{equation~\ref{#1}}

\def\plaineqref#1{\ref{#1}}

\def\1{\bm{1}}

\DeclareMathAlphabet{\mathsfit}{\encodingdefault}{\sfdefault}{m}{sl}
\SetMathAlphabet{\mathsfit}{bold}{\encodingdefault}{\sfdefault}{bx}{n}

\usepackage{amsmath,amssymb}
\usepackage{booktabs}
\usepackage{longtable}
\usepackage{graphicx}
\usepackage{wrapfig}
\usepackage{microtype}
\usepackage[table]{xcolor}
\usepackage{hyperref}
\usepackage{url}

\newcommand{\traitspace}{\mathbb{B}}                 \newcommand{\resource}{B}                             \newcommand{\pos}[1]{x_{#1}}                          \newcommand{\allpos}{\mathbf{x}}                      \newcommand{\kernel}{f_\kappa}                        \newcommand{\share}[1]{G_{#1}}                        \newcommand{\util}[1]{u_{#1}}                         \newcommand{\reach}{\sigma}                           \newcommand{\reachstar}{\sigma^{*}}                   \newcommand{\nspec}{N}                                \newcommand{\ndim}{L}                                 

\newcommand{\depth}{D_\kappa}                         \newcommand{\curv}{K}                                 \newcommand{\scatter}{C}                              \newcommand{\score}{s}                                \newcommand{\keff}{K_{\mathrm{eff}}}

\newcommand{\expectB}[1]{\mathbb{E}_{\resource}\!\left[#1\right]}

\newcommand{\method}{\textsc{InflEns}}
\definecolor{oursrow}{RGB}{222,235,250}
\newcommand{\oursrow}{\rowcolor{oursrow}}

\newif\iffinalcopy
\finalcopyfalse

 \graphicspath{{figures/}}

\hypersetup{
colorlinks=true, linkcolor=black, citecolor=black, urlcolor=blue, pdftitle={Game-Theoretic Data Allocation for Expert Specialization and Routing}, pdfauthor={Mark Lovett, Denis Lim, Soroush Vosoughi}, }

\title{Game-Theoretic Data Allocation for Expert Specialization and Routing}

\author{
Mark Lovett\thanks{Correspondence: \texttt{mark.lovett.gr@dartmouth.edu}} \\ Dartmouth College
  \And
Denis Lim \\ Independent Researcher
  \And
Soroush Vosoughi \\ Dartmouth College }

\iclrfinalcopy

\begin{document}

\maketitle
\lhead{Preprint}

\begin{abstract}

Can the allocation used to train experts also predict which expert will perform best on a prompt? Such a connection would make the intended division of labor inspectable before training and allow the allocation itself to guide inference. We introduce \method{}, a mixture of LoRA experts whose training weights and inference routes come from the same location game in a fixed prompt-trait space. Players choose positions to maximize their expected shares of the prompt distribution, without using expert losses. We utilize the game's kernel-general theoretical criterion for when co-located positions become locally unstable to separation. We evaluate \method{} on a corpus assembled from seven safety-related benchmarks, which demonstrates that the highest-share expert has the lowest held-out response negative log-likelihood (NLL) on approximately 90\% of prompts across four kernels under a sharpened simplex representation, compared with approximately 15\% after permuting allocations across prompts. We demonstrate that the method performs comparably to published learned gates without having to train an explicit gate on the experts. Because the router is fixed, we can trace harmful behavior to a single expert and retrain only that expert, which lowers HarmBench attack success from 0.51 to 0.02 while keeping instruction following. The game also gives us several controls over how the experts specialize, including the kernel choice and its parameters, the trait-space geometry, and the number of experts. Importantly, \method{} is a general framework that can be extended to any encoder and data distribution.

\end{abstract}

\section{Introduction}
\label{sec:intro}

In many mixture-of-experts systems, expert specialization develops as the gate and experts are optimized together \citep{shazeer2017moe}. We ask whether this division of labor can be specified through a training allocation that also predicts where each expert will be useful. An explicit allocation would let a designer inspect which experts receive training signal from particular kinds of prompts, such as privacy-sensitive requests or attempted jailbreaks, before training the adapters. It could also supply a rule for selecting among the trained experts after training has been conducted. 

We introduce \method{}, a method for training experts in a trait space using game dynamics, and we demonstrate \method{}'s applicability with a mixture of LoRA experts \citep{hu2022lora}. To start, \method{} needs to have a fixed representation that maps data, prompts in this case, to trait vectors, which in turn builds a resource space that players can compete over. With the resource representation in hand, players choose positions in the resource trait space, and an influence kernel centered on the players' positions determines the share of each data point's (here, each prompt's or prompt response pair's) learning signal that a player receives relative to its competitors. Each player aims to maximize its expected share of the training signal. Note that this differs from methods trained on loss, because the players are not explicitly trained to minimize their loss, only to maximize how much training signal they receive. Because the shares of each data point's signal sum to one, the game is constant-sum, so no data point's signal goes to waste and one player's gain comes at another's expense. In our example setting with prompts as data and LoRA adapters as players, the players' shares weight their associated adapters' training losses and determine their inference allocation. Before we start training we solve the position game numerically, so that players start near equilibrium, but we show other starts as well. We then allow positions to adapt through a gradient ascent algorithm with game-utility gradients as the guiding gradient; this allows our system to move toward stable equilibria as more of the data distribution is revealed, and organizes the experts into predictable positions within their trait space. Thus, \method{} results in the predictable routing of training signal to experts and leads to predictable emergence of experts and ultimately requires no training of a gate on losses. 
 
\begin{figure}[t]
  \centering
  \includegraphics[width=\linewidth]{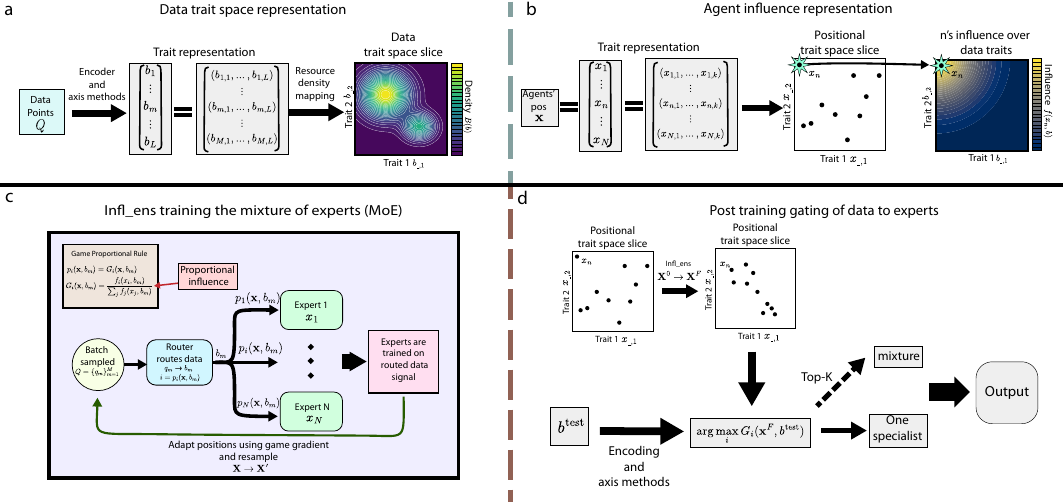}
  \caption{\method{} makes the allocation of training data explicit through a location game, the influencer's game, which requires a resource distribution for players to compete over. \textbf{a}~We construct the resource distribution by encoding text data and projecting it onto trait axes (Appendix~\ref{app:trait-geometry}), giving a density over the trait space. \textbf{b}~Each player holds a position in their trait space (the same as the resource space in this work), and an influence kernel centered on that position determines how much of each data point's training signal the player receives. \textbf{c}~During training, each sampled batch is routed to the experts in proportion to their influence shares, the experts train on their weighted signal, and the positions are then updated by the game gradient before the next batch is sampled. \textbf{d}~After training, the experts have settled into positions in the trait space, and a new prompt is routed by encoding it and selecting the expert with the largest share, or a top-$k$ mixture, without a learned gate.}
  \label{fig:intro-overview}
  \vspace{-0.8em}
\end{figure}

Existing clustering-based methods also use data geometry to organize expert training and inference \citep{gururangan2023cbtm,gou2026mocle}. A distinguishing factor of \method{} is that positions follow from individual competitive payoffs derived from the influencer's game framework \citep{lovett2026influencer}. In \method{} we utilize the stability properties of the influencer's game, which provides a framework for analyzing and predicting agents' behavior. The influencer's game provides a symmetric stability criterion that applies across log-concave influence kernels and gives an explicit boundary determined by player count, kernel geometry, and prompt-trait covariance. The influencer's game's boundary provides us a reference scale for kernel reach (the curvature of the kernel) and identifies when co-location becomes unstable and gives way to separation of players. We then test whether the induced allocation predicts the relative response loss of the trained experts.

In the setting of this paper, \method{} is evaluated against other training methods on a corpus of prompts assembled from seven safety-related benchmarks, from which we build a fixed, benchmark-informed trait representation. This provides a resource distribution on which players apply the influencer's game's dynamics. For the prinicple experiment the trait map and resource distribution use the full prompt population, while test examples are excluded from adapter optimization and gate fitting. The trait-space representation and learend axises of the data corpus can vary, and we deomonstrate that the method is not limited to the specific trait-space representation.

We also run separate experiments to test whether \method{} improves predictive performance. In the principal seven-adapter configuration, token-level probability mixing improves NLL by 0.105 nats over a pooled adapter, and with the trained experts held fixed, game-weighted mixing outperforms uniform weighting, showing that the prompt-dependent allocation contributes to the gain. Across four data splits, the game allocation also achieves lower expected NLL than validation-trained linear gates when sampling one expert, while completed-sequence probability mixtures perform similarly.

The game gives us a way to study expert organization through choices about training data, competitive positions, and influence, which makes the division of training data both inspectable and testable through its predictions about expert performance.

\subsection{Related work}

\paragraph{Sparse mixtures and routing.}
Pre-existing adaptive mixtures and sparse Transformer MoEs conventionally learn an input-dependent gate jointly with their experts \citep{jacobs1991adaptive,shazeer2017moe,fedus2022switch}. Alternative assignment schemes show that such joint learning is not always necessary: token hashing and stochastic expert selection can rival learned routers in the settings studied, whereas StableMoE learns a router in an initial stage and then freezes it to limit assignment drift \citep{roller2021hash,zuo2022thor,dai2022stablemoe}. Capacity balancing introduces a separate trade-off; for instance, BASE guarantees equal per-batch expert loads through constrained assignment \citep{lewis2021base}, potentially forcing certain tokens to be routed to sub-optimal experts simply to satisfy the strict uniform distribution constraints. Conversely, computing load-balancing losses within each micro-batch can spread domain-specific sequences across experts and inhibit domain specialization \citep{qiu2025demons}. Studies of existing trained MoE routers have found that what an expert specializes in depends on the method; for instance, OpenMoE routes mostly by token identity \citep{xue2024openmoe}, while Mixtral's experts show little topical separation but Mixtral groups tokens by syntax and by position in the sequence \citep{jiang2024mixtral}, and OLMoE shows some specialization by domain and vocabulary \citep{muennighoff2025olmoe}. A recent preprint further relates expert-use similarity to hidden-state geometry \citep{wang2026myth}. \method{} instead defines prompt allocation through the influencer's game in a trait space; thus, a key defining feature of \method{} is that its allocations depend on game-derived positions rather than adapter performance. A benefit of \method{}'s game structure is that the initial allocation and the local stability of co-located positions can be analyzed before fine-tuning, and the experts' coordinates determine their subsequent routes rather than a gate learned on their losses.

\paragraph{Expert libraries, composition, and post-hoc routing.}
Modular systems are another line of work, and they mainly differ in when specialization is decided and how the experts are recombined afterward. BTM trains full-model branches on domains defined by metadata and then combines the branches' predictions using a domain posterior derived from their likelihoods \citep{li2022btm}. c-BTM builds on this by replacing the metadata with clusters of the training corpus, and the cluster centers are then reused at inference to weight each expert \citep{gururangan2023cbtm}. BTX takes a different route, first training domain experts and then converting them into an MoE, where a token-level gate is learned on top of them \citep{sukhbaatar2024btx}. At the scale of adapters, MoCLE follows a similar idea to c-BTM, clustering instruction embeddings and activating LoRA experts based on how similar an input is to each cluster \citep{gou2026mocle}. A separate group of methods starts only after the adapters have been trained. AdapterSoup averages the weights of a selected set of adapters \citep{chronopoulou2023adaptersoup}, MoLE and PHATGOOSE learn a gate over the trained adapters \citep{wu2024mole,muqeeth2024phatgoose}, Arrow builds routing prototypes directly from the adapter weights so that no further training is needed \citep{ostapenko2024arrow}, and LoraRetriever learns which adapters to retrieve for a given input \citep{zhao2024loraretriever}. At the level of whole models, RouteLLM learns from preference data which model should answer a prompt \citep{ong2025routellm}, and recent work compares routing, merging, and ensembling of parameter-efficient experts head to head \citep{lotfi2026tradeoffs}. Of these, c-BTM and MoCLE are the closest to \method{}, since in both the geometry of the clusters shapes what each expert trains on and how it is used at inference. The distinguishing factor of \method{} is that the positions organizing the experts come from the influencer's game rather than from clustering, and because these positions are solved before any adapter is trained, their shares can route prompts without needing a gate fitted on the experts' losses.

\paragraph{Game-theoretic and competitive mixtures.}
Game theory has also been applied to multi-expert learning, and prior work mainly differs in who the players are and what they control. In Nash-MTL the players are tasks, which bargain over a shared parameter update \citep{navon2022nashmtl}, while mean-field and congestion formulations instead treat tokens as the players, trading off routing quality against the load placed on each expert \citep{sen2026meanfield,mouzouni2026three}. Strategic-predictor models take yet another view, letting learners compete for the data points they can serve accurately \citep{benporat2019regression}. Within MoEs specifically, Mixture of Bidders routes batches through VCG auctions \citep{vyas2025mob}, NAMEx uses bargaining to set the weights for merging experts \citep{nguyen2026namex}, and a gate-free vision MoE uses each expert's marginal contribution to decide how much it participates \citep{agarap2026gatefree}. CompeteSMoE trains a router to reproduce the assignments of whichever experts respond most strongly \citep{nguyen2025competesmoe}, and a related model of softmax routing finds a pitchfork bifurcation, but in the load placed on each expert rather than in the experts' positions \citep{kiselev2026minimal}. In \method{} the players are the experts themselves, and each one chooses a position in a fixed trait space in order to earn a share of the training signal. We solve the influencer's game numerically to give the players their starting positions and use its local stability analysis to predict when those positions separate, and the resulting shares then weight adapter training and route inference.
 \section{The Influencer's Game}
\label{sec:game}
We use the influencer's game of \citet{lovett2026influencer}. Let a data point's frozen trait vector be $b\in\traitspace$, drawn from a resource distribution $\resource$. Player $i\in\{1,\ldots,\nspec\}$ chooses position $\pos{i}\in\traitspace$. Then for a positive influence kernel $\kernel(x,b;\reach)$ with parameter $\sigma$, a player's share and expected payoff are
\begin{equation}
    \share{i}(\allpos,b)=\frac{\kernel(\pos{i},b;\reach)}
     {\sum_{j=1}^{\nspec}\kernel(\pos{j},b;\reach)},\qquad
    \util{i}(\allpos)=\expectB{\share{i}(\allpos,b)}.
    \label{eq:allocation}
\end{equation}

For every resource point $\sum_i\share{i}=1$, hence $\sum_i\util{i}=1$ at every configuration, indicating the game is constant-sum. Importantly, being constant-sum implies that one player's gain comes at the expense of the other player(s). The constant-sum nature means that specialists will be competing for the same resource signal and are therefore encouraged to explore new niches when a niche is crowded and their influence kernel is not sufficiently wide. Importantly, the utility of the player is based on their training signal share, and their accuracy is an emergent property of the training on that signal. In turn, this means that a player's received training signal is tractable through the history of the game, allowing us to predict specialists without explicitly testing their accuracy during training. 

\subsection{Equilibrium and instability}
\label{sec:game:stability}
We start by assuming that players are at a symmetric profile $x_1=\cdots=x_\nspec=x^\circ$, meaning they share a position in the trait space. We let the first derivative of a player's log-influence at the trait point $\score(b)$ be $\score(b)=\nabla_x\log\kernel(x^\circ,b;\reach)$. Then an interior symmetric stationary point obeys $\expectB{\score(b)}=0$. Define restoring log-curvature and score scatter by
\begin{equation}
    \curv=-\expectB{\nabla_x^2\log\kernel(x^\circ,b;\reach)},
    \qquad \scatter=\expectB{\score(b)\score(b)^\top}.
\end{equation}
When $\curv$ is positive definite on the relevant tangent space, the antisymmetric, player-separating mode has instability depth
\begin{equation}
    \depth=\frac{\nspec-2}{\nspec-1}\,
    \lambda_{\max}\!\left(\curv^{-1/2}\scatter\curv^{-1/2}\right).
    \label{eq:depth}
\end{equation}
For $\depth<1$ the symmetric profile is locally stable, while $\depth>1$ gives a player-separating linear instability. Here $\depth=1$ is the stability boundary and, under a transversal parameter crossing, the first local bifurcation from the symmetric equlibirium. Importantly, the log-concavity of a kernel indicates there is always a kernel parameter regime where the symmetric equilibrium is stable and unique. Additionally, for two effective players, the prefactor vanishes, so there is no separating-mode bifurcation and the symmetric equilibrium is always locally stable. Appendix~\ref{app:derivations} derives these claims from Equation~\plaineqref{eq:allocation}. 

We study Gaussian, hyperbolic, and exponential-power kernels whose parameters determine the kernel's spread from its mode (standard deviation, etc.), as well as kernels whose parameters determine the kernel's shape, including the simplex-native Dirichlet kernel and a coordinate-wise product-Beta kernel. We denote a kernel's reach by $\sigma$, where Gaussian log influence scales as $\reach^{-2}$, hyperbolic as $\reach^{-1}$, and exponential-power as $\reach^{-\beta}$; the shape kernels place inverse reach inside their density parameters and normalizers. To simplify comparisons across kernel types, we compare kernels by their depth in Equation~\plaineqref{eq:depth} and by their allocation sharpness (Section~\ref{sec:design}); Appendix~\ref{app:kernels} gives every implemented closed form.
 \section{Methods}
\label{sec:exp:setup}

The game router needs to have a trait assigned to every data point so that experts have a space to compete within for training signal. Beyond the requirement of a trait space, other properties of the data distribution, such as dimensionality or labels, do not limit the applicability of \method{}. Naturally, many empirical data sets have high-dimensional trait spaces when using raw embeddings, such as the 4096-dimensional embeddings of Qwen3-Embedding-8B \citep{qwen2025embedding}, which likely leads to very sparse resource distributions with limited data sets. Resource sparsity in 

\begin{wrapfigure}[40]{r}{0.3\textwidth}
  \vspace{-1em}
  \centering
  \includegraphics[width=\linewidth,trim=12 0 248 277,clip]{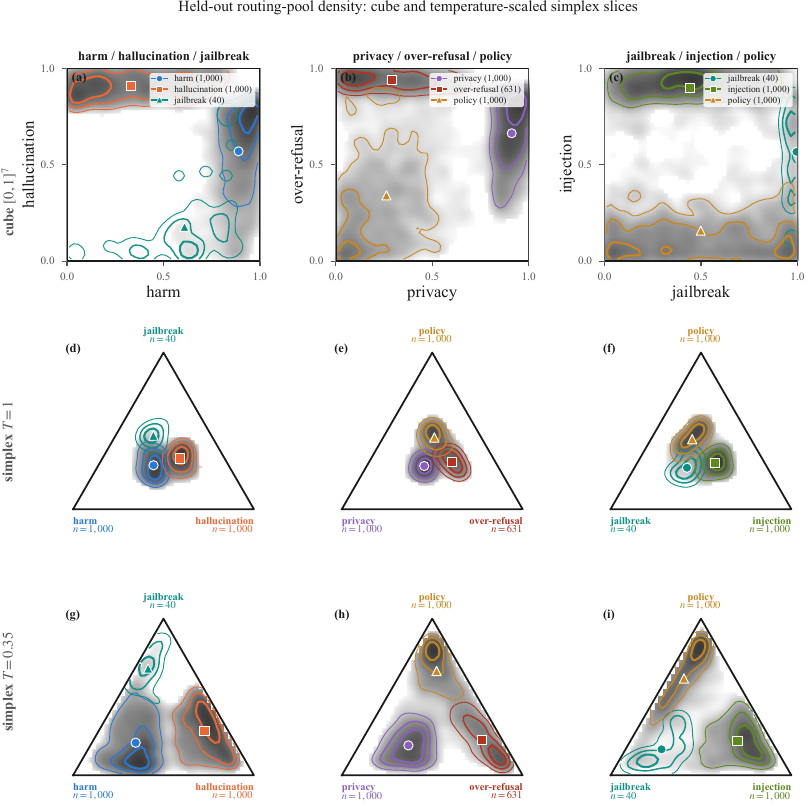}\\[0.3em]
  \includegraphics[width=\linewidth,trim=139 0 122 277,clip]{fig_research_trait_slices.pdf}\\[0.3em]
  \includegraphics[width=\linewidth,trim=262 0 0 277,clip]{fig_research_trait_slices.pdf}
  \caption{\textbf{Trait Space example}; held-out prompt density in three conditional simplex views at $T=0.35$. Grey is the combined density; colored contours mark benchmark origin. Panels (g)--(i) are the bottom row of Figure~\ref{fig:trait-slices}, which also gives the cube and $T=1$ views.}
  \label{fig:trait-simplex}
  \vspace{-1em}
\end{wrapfigure}

itself does not limit the effectiveness of \method{} but can make it more difficult to demonstrate the nuances of its impacts. To address this, we transform the 4096-dimensional embeddings of Qwen3-Embedding-8B to seven axes defined by safety-relevant data sets; Appendix~\ref{app:label-free} replaces them with axes learned without labels. We generate a mixed dataset from seven benchmarks on distinct safety-relevant properties: harm, hallucination, jailbreak, privacy, over-refusal, policy violation, and prompt injection (Appendix~\ref{app:data-provenance}). Each safety-relevant property then defines an axis label that we use for data points. The loss and the reported NLL cover the full chat text, and four of the seven sources contribute prompts only (47\% of training records), so on those the models are trained and scored on the prompts themselves; about 10\% of training targets are unsafe responses and 9\% hallucinated answers (Table~\ref{tab:corpus-composition}).

To generate the trait space, we use a frozen, 4-bit quantized Qwen3-Embedding-8B encoder throughout. For each axis we fit a shrinkage Fisher discriminant direction in the embedding space, using the benchmark’s own labels (for example, harmful versus harmless responses) blended with a direction that separates that benchmark’s prompts from the rest. Each new direction is orthogonalized against the directions already fitted, and each coordinate is residualized against the earlier coordinates, so that overlapping benchmarks do not collapse onto a single axis. The resulting scores are mapped to $[0,1]$ by a per-axis empirical CDF, so each coordinate is a percentile within the corpus and the seven marginals are uniform over the corpus by construction. The simplex variant applies a softmax at temperature $T=0.35$ to these coordinates. Both the cube and the simplex are shown in Figure~\ref{fig:trait-slices} in Appendix~\ref{app:trait-geometry}, where the construction is given in full. The 28,355 prompts of the combined corpus, mapped into this space, form the resource distribution that the experts compete over for training signal.

The principal experiment places fourteen clones of the base model at seven co-located pairs of positions and trains one rank-16 LoRA adapter per pair. The pairing of the clones is motivated by two game properties: a symmetric two-player game with positive log-concave influence kernels has no player-separating instability (Section~\ref{sec:game:stability}), and duplicating every solved location into an equal co-located pair leaves the effective allocation exactly unchanged (Appendix~\ref{app:derivations}). In effect, this implies that the pairs behave as one agent when updating their positions, with the added benefit of increased diversity at a lower depth than individual initialization. The pair is therefore the routing and adapter-training unit: its members' shares sum to the allocation a single agent would receive at that position in the seven-player game, but at a much lower depth. We keep the even split between members only in the position dynamics, where each clone follows its own utility gradient in the fourteen-player game. Equal co-located clones receive identical updates, so pairing is an invariant of the update rather than an explicit copy operation (Appendix~\ref{app:position-update}). Solving the eight-player position game directly shows the same invariance, as players started in four co-located pairs remain paired at every reach below $0.7\reachstar$ (Figure~\ref{fig:group-tree}).

In each experiment, the pipeline completes 12 rounds, consisting of the following steps: it projects a training batch to traits, computes pair shares by Equation~\plaineqref{eq:allocation}, and assigns each prompt to its top-$k$ pairs, with retained shares renormalized as loss weights. The dense $k=N$ arm retains all $N$ pairs, whereas $k=1$ is a hard allocation. In the dynamic-position arm, after allocation, each clone takes a simultaneous projected ascent step on its game payoff, using the score of the chosen influence kernel over the whole round batch. A fixed-position control disables every position-update path while leaving assignment and adapter training otherwise unchanged. Appendix~\ref{app:routing-ablation} separately ablates fan-out, sampling, and loss weighting, showing why the main experiments retain the full share-weighted allocation.

At evaluation, final pair positions yield $G_p(b)$ without fitting a gate to held-out expert losses. We distinguish expected NLL $\sum_pG_p(b)\ell_p(b)$ (the loss of sampling one pair), argmax commitment, and the NLL of mixing pair probabilities over a completed sequence, which can exploit probability averaging that a single-expert decision cannot. In each experiment, we use a hindsight best-expert oracle to supply an upper bound on recoverable routing headroom. To show how a single model compares, a pooled comparator trains on the corresponding union of routed batches, matching the training signal the seven experts receive in aggregate. Therefore, an expert and pooled adapter each receive the same 1,248 optimizer steps. 

The 28,355 records (Table~\ref{tab:resource-data}) are partitioned independently within each source using a persisted seeded manifest: 19,848 train, 2,836 validation, and 5,671 test records (a 70/10/20 split). Every arm reuses those indices.We fit the trait axes, quantile normalization, and resource KDE to the full 28,355-record population before any adapters are trained; this resource KDE is only used for calculating the theoretical equilibrium of the distributions. ``Test'' therefore means held out from adapter optimization, gate fitting, and expert-loss feedback, but not from resource-map estimation. Refitting all three on the training partition only lowers the routed single-expert gain by 0.0091 nats on three splits without changing expert quality (Appendix~\ref{app:inductive}). The principal arm starts from Qwen2.5-1.5B-Instruct \citep{qwen2024qwen25}, trains rank-16 LoRA adapters for twelve rounds, uses soft routing with weights equal to the proportion of influence over data (Appendix~\ref{app:routing-ablation}), and scores a 5,671-prompt held-out pool. Appendix~\ref{app:model-ablations} ablates the base model's family and size.
 \section{Results}
\label{sec:results}
\label{sec:design}
\label{sec:experiments}

\subsection{Game Controls: Equilibrium Positioning, Influence Kernel, and Resource Geometry}
\label{sec:design:regime}

\begin{figure}
  \vspace{-1em}
  \centering
  \includegraphics[width=.65\linewidth]{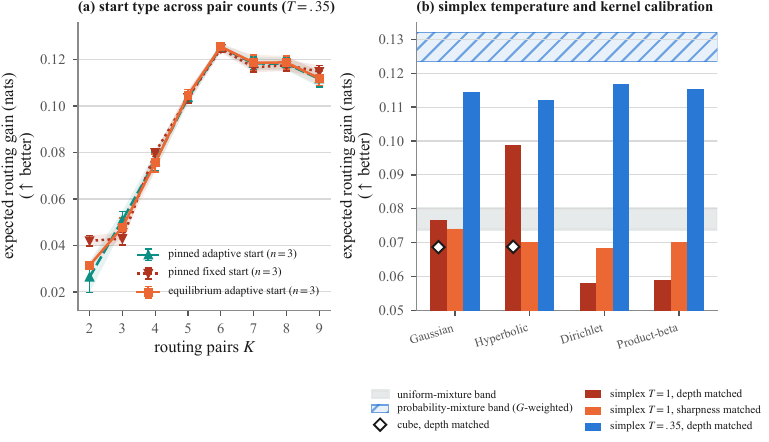}
\caption{The starting configuration changes little once the game is depth matched on the sharpened simplex, whereas induced allocation sharpness matters more than kernel identity. (a) Expected routing gain is pooled-generalist NLL minus the share-weighted NLL of sampling one pair (higher is better), for the $T=0.35$ Dirichlet game at two to nine pairs under three starts; bands are one standard deviation across three data splits. (b) Matching bifurcation depth leaves a large gain spread across four kernels (dark red); matching prompt-level sharpness narrows it (orange), and sharpening the simplex projection to $T=0.35$ (dispersed) raises and nearly equalizes the gains (blue). Hollow diamonds are cube controls.}
  \label{fig:init-kernels}
  \vspace{-1em}
\end{figure}

Figure~\ref{fig:init-kernels}a shows \method{} trained on the $T=0.35$ Dirichlet game with two to nine pairs, depth matched to the seven-pair arm, on three data splits under the pinned adaptive, pinned fixed and equilibrium adaptive starts compared in Section~\ref{sec:kernel-params}. Gain peaks at 0.125 nats at six pairs under all three, and the starts agree within their split spread from four pairs on. At this reach, however, the exact equilibrium lies only 0.005--0.020 from the unsolved start (neighbouring pairs are 0.10--0.24 apart), so the comparison cannot show whether a distant equilibrium start helps (Figure~\ref{fig:radar-pairs}).

In Figure~\ref{fig:init-kernels}b, four kernels and three trait-space geometries show that trait-space geometry shifts how much kernel identity matters. Matched on bifurcation depth on the simplex at $T=1$ (dense), expected gain spans 0.0407 nats across the four kernels and tracks their sharpness almost exactly, with the sharpest kernel best and the most diffuse worst. Matching sharpness by shifting kernel parameters collapses that spread to 0.0055 nats. Meanwhile, sharpening the simplex projection to $T=0.35$ (dispersed) lifts every kernel to roughly 0.115 nats with a spread of 0.0046, effectively eliminating any difference between the kernels when depth matched. Table~\ref{tab:kernel-interpretability} shows the same collapse in how well the allocation predicts expert quality: top-1 agreement spans 82.6--87.3\% across kernels at $T=1$ and 89.2--90.8\% at $T=0.35$, against a shuffled-allocation null of about 15\%. This indicates that kernel choice and resource distribution impact the predictive power of the game router. 

Importantly, in Table~\ref{tab:kernel-interpretability} the game router ranks the best expert first about six times as often as chance, indicating that specialization follows from \method{}'s allocation. The simplex geometries also show higher router--oracle agreement and mixture capture than the cube, but a bidirectional frozen-expert control shows that each expert family prefers the geometry in which it trained, so experts trained in one geometry do not transfer to another (Appendix~\ref{app:trait-geometry}).

\begin{table}[!t]
\centering
\footnotesize
\setlength{\tabcolsep}{4pt}
\caption{How well the game router predicts expert quality, per influence kernel and simplex temperature $T$, which sets how concentrated the resource distribution is. Each cell is the mean $\pm$ sample standard deviation over three training seeds of the seven-pair run; the seeds share one data split. Higher is better for every measure. Top 1, top 2 and top 3 are the percentages of held-out prompts whose best expert receives the largest share, one of the two largest, or one of the three largest; each null shuffles allocations across prompts. Spearman is the mean within-prompt rank correlation between allocation and expert quality.}
\label{tab:kernel-interpretability}
\begin{tabular}{llrrrrrrrr}
\toprule
 & & \multicolumn{2}{c}{Top 1 (\%)} & \multicolumn{2}{c}{Top 2 (\%)} & \multicolumn{2}{c}{Top 3 (\%)} & \multicolumn{2}{c}{Spearman} \\
\cmidrule(lr){3-4} \cmidrule(lr){5-6} \cmidrule(lr){7-8} \cmidrule(lr){9-10}
$T$ & Kernel & obs. & null & obs. & null & obs. & null & obs. & null \\
\midrule
1.00 (dense) & Gaussian & 86.3{\scriptsize\,$\pm$\,0.6} & 15.1 & 95.2{\scriptsize\,$\pm$\,0.3} & 26.4 & 97.7{\scriptsize\,$\pm$\,0.3} & 39.7 & 0.719{\scriptsize\,$\pm$\,0.014} & 0.204 \\
 & Hyperbolic & 87.1{\scriptsize\,$\pm$\,0.3} & 15.0 & 96.4{\scriptsize\,$\pm$\,0.2} & 27.3 & 98.4{\scriptsize\,$\pm$\,0.1} & 41.3 & 0.641{\scriptsize\,$\pm$\,0.008} & 0.146 \\
 & Dirichlet & 82.9{\scriptsize\,$\pm$\,1.3} & 15.2 & 92.6{\scriptsize\,$\pm$\,0.5} & 26.2 & 95.8{\scriptsize\,$\pm$\,0.4} & 39.2 & 0.729{\scriptsize\,$\pm$\,0.010} & 0.206 \\
 & Product-beta & 83.2{\scriptsize\,$\pm$\,1.8} & 15.2 & 93.0{\scriptsize\,$\pm$\,0.6} & 26.2 & 96.1{\scriptsize\,$\pm$\,0.5} & 39.3 & 0.725{\scriptsize\,$\pm$\,0.015} & 0.206 \\
\addlinespace
0.35 (dispersed) & Gaussian & 90.5{\scriptsize\,$\pm$\,0.4} & 15.4 & 96.7{\scriptsize\,$\pm$\,0.1} & 28.7 & 98.2{\scriptsize\,$\pm$\,0.2} & 44.1 & 0.518{\scriptsize\,$\pm$\,0.020} & 0.069 \\
 & Hyperbolic & 89.3{\scriptsize\,$\pm$\,0.5} & 15.2 & 96.6{\scriptsize\,$\pm$\,0.1} & 27.8 & 98.4{\scriptsize\,$\pm$\,0.1} & 42.7 & 0.565{\scriptsize\,$\pm$\,0.016} & 0.108 \\
 & Dirichlet & 90.1{\scriptsize\,$\pm$\,0.3} & 15.2 & 96.6{\scriptsize\,$\pm$\,0.3} & 28.1 & 98.0{\scriptsize\,$\pm$\,0.2} & 42.8 & 0.469{\scriptsize\,$\pm$\,0.007} & 0.044 \\
 \oursrow& Product-beta & 90.5{\scriptsize\,$\pm$\,0.3} & 15.3 & 96.6{\scriptsize\,$\pm$\,0.3} & 28.2 & 98.1{\scriptsize\,$\pm$\,0.2} & 43.0 & 0.475{\scriptsize\,$\pm$\,0.007} & 0.044 \\
\bottomrule
\end{tabular}
\end{table}
\subsection{Kernel Parameters}
\label{sec:kernel-params}

\begin{wrapfigure}[17]{l}{0.5\textwidth}
  \vspace{-1em}
  \centering
  \includegraphics[width=\linewidth]{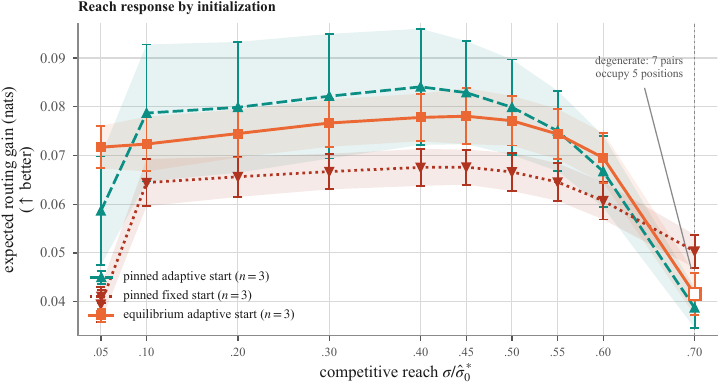}
\caption{The equilibrium adaptive start stabilizes the reach response without changing mean gain; freezing pinned positions costs gain without adding regularity. Each point is a within-data-seed contrast against its pooled generalist; bands are sample standard deviations across three splits per condition.}
  \label{fig:init-reach}
  \vspace{-1em}
\end{wrapfigure}

Kernel parameters control how broadly an expert competes for training signal and can determine the depth of the influence kernel past its bifurcation threshold. To test this effect on \method{}, in Figure~\ref{fig:init-reach} we limit to seven pairs, use the first local bifurcation $\reachstar$ as a reference scale for a spherical multivariate Gaussian influence kernel's reach parameter $\sigma$, and then sweep the reach and measure the resulting routing response. We compare three conditions: an equilibrium adaptive start, which begins at the solved empirical-game equilibrium and tends to stay near it; a pinned adaptive start, which begins every reach on a data split from the same unsolved configuration; and a pinned fixed condition, identical but with positions never updated (Appendix~\ref{app:initialization}).

Initialization acts on the shape of the reach response and leaves its level unchanged. Mean per-split roughness (excluding the degenerate 0.70 endpoint) is 0.483 for the equilibrium adaptive start and 1.266 for the pinned start over three data splits each, and the per-split ranges do not overlap (0.43--0.53 against 1.07--1.45).

Both starts peak at reach 0.40 or 0.45, so a shared start fixes where the response peaks, but the equilibrium adaptive start is 2.6 times smoother and makes the choice of which configuration to pin less arbitrary. Mean gain does not separate the two adaptive starts, at 0.0764 nats for the pinned and 0.0747 nats for the equilibrium adaptive start, and the paired difference, $-0.0018 \pm 0.0082$ nats across splits, is small next to its own spread (Figure~\ref{fig:init-reach}).

The pinned start runs had not settled by round 11: they closed 43\% of their distance to the equilibrium adaptive runs on average, and continuing the position update past training carries only 15 of the 27 non-degenerate reach--split cells to the same fixed point (Appendix~\ref{app:initialization}), indicating there is some choice of equilibria that needs to be made.

\subsection{Routing evidence and controls}
\label{sec:exp:routing}
With the seven-pair expert family fixed, we first try to determine how much of the gain comes from how the experts' outputs are combined. Under the standard token-level estimator, the routed mixture gains 0.1049 nats over pooling all of the data to one generalist model. Additionally, the completed-sequence estimator we use for the same-expert nulls gives a slightly more optimistic 0.1093--0.1095 nats. On that same estimator, mixing the experts uniformly still retains a 0.0661-nat gain, whereas shuffling the expert identities loses 0.0671 nats and uniform random hard routing loses 0.4144, which shows that the game's allocation carries information about which expert suits which prompt (Appendix~\ref{app:routing-ablation}, Figure~\ref{fig:router-nulls}). In these nulls we only change the combiner, and the same appendix shows the impact of changing the way the adapters are trained. Throughout, we report the dense top-$k=N$ case as our default, where $N$ is the number of pairs and every pair contributes to the mixture, but other methods are shown in Appendix~\ref{app:routing-ablation}.

To compare \method{} to a learned gate, we first freeze all adapters and fit linear prompt-conditioned gates on the validation expert losses. Across four data splits, the game allocation beats both the full-NLL soft-target gate and a best-expert classification gate by 0.071--0.090 nats when a single expert is committed to per prompt. Under full probability mixing, the sign divides two-to-two across splits and the largest difference is only 0.0017 nats, which is at the measured run-to-run noise floor. These results support that a gate-free allocation is sufficient for hard routing to experts.

\begin{figure}[t]
  \centering
  \includegraphics[width=.75\linewidth]{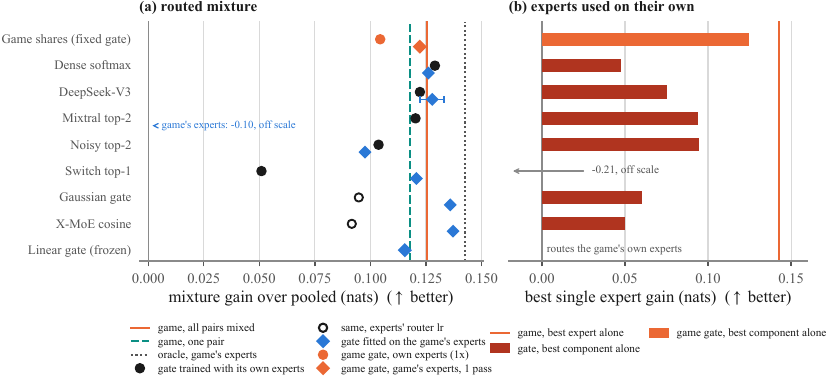}
  \caption{Published gating rules against the game on the $T=0.35$ simplex (seed-0 split, one run per arm). (a) Routed-mixture NLL as gain over the pooled generalist (higher is better), token level throughout. Filled markers are gates trained jointly at router learning rate $10^{-2}$, hollow markers at the experts' rate, and diamonds are gates fitted post hoc on the game's frozen experts; the top row fixes the gate to the game's shares, and the dotted line is the hindsight-best expert. (b) The per-prompt best single component of each mixture against the game's best single expert. Runs at the experts' router learning rate are compared in Appendix~\ref{app:gates}.}
  \label{fig:gates}
  \vspace{-0.8em}
\end{figure}

We also compare \method{} with routers trained end to end. We implement five published gates (dense softmax, noisy top-$k$, Switch, Mixtral and DeepSeek-V3) and two distance gates (Gaussian and X-MoE cosine). Each gate is trained jointly with seven LoRA experts on the game's batches, using the same $T=0.35$ trait coordinates as input (Figure~\ref{fig:gates}). We train the five published gates with the router learning rate raised to $10^{-2}$, because at the experts' learning rate a prompt-level router barely moves in 1{,}248 steps (Appendix~\ref{app:gates}). With that correction, the dense softmax gate (1.8864) is within 0.0035 nats of the game's token-level mixture over all pairs (1.8899), and DeepSeek-V3 and Mixtral, which route each prompt to two experts, trail the game's top-2 mixture (1.8901) by 0.003 and 0.005 nats, all within run-to-run noise. The difference between the methods lies in the experts themselves: the best single component of a jointly trained mixture gains only 0.05--0.09 nats over pooling, against 0.143 for the game's best expert, so the game's experts can be served on their own while the gates' components only work when blended.

The game's allocation is not the only way to route its own experts. When the frozen linear soft-target gate routes the game's experts, it scores 1.8999 under the same token-level estimator, 0.0101 nats behind the game's shares. When the published gates are instead fitted directly on the game's frozen experts, the distance gates route them better than the game's shares, with X-MoE cosine scoring 1.8782 and the Gaussian gate 1.8795 (0.0116 and 0.0104 nats ahead), while dense softmax and DeepSeek-V3 match the game and noisy top-$k$, Switch and Mixtral trail it (Appendix~\ref{app:gates}). The game's allocation is therefore not the best possible router for its own experts, but it is a good one, and a distinguishing factor is that it is fixed before training rather than learned on the experts' losses.

Finally, we test whether the game's allocation still produces experts when it is used as the gate of a single jointly trained mixture. Fixing the gate to the game's shares costs the same as training a learned gate and still yields experts: its components alone reach 0.125 nats over pooling at the oracle, against 0.05--0.09 for the learned gates, while the blended network scores 1.9110 (Appendix~\ref{app:gates}). We can also blend the game's separately trained experts into one network under the same gate, which needs no further training and scores 1.8932, within 0.0033 nats of mixing their outputs, so the game's mixture does not require seven forward passes at inference. The game also needs no router tuning, although it uses seven times the training compute for the same number of updates per adapter.

\subsection{Additional experiments}
\label{sec:exp:behavioral}

Experimenting with Qwen2.5 model sizes of 1.5B, 3B, 7B, 14B, and 32B shows that each size improves over its own pooled comparator while oracle-headroom capture declines, and the allocation effects also transfer to two other model families, Gemma and Llama (Appendix~\ref{app:model-ablations}).

Additionally, \method{} compares favorably within a ten-family baseline audit, including a frozen gate, same-expert nulls, balanced c-BTM, BASE-style assignment, MoLoRA, KnOTS, and a pooled-seed deep ensemble (Appendix~\ref{app:baselines}). In the most directly comparable of these, on the same seed-0 split and estimator, the game mixture nearly doubles c-BTM's gain over pooling (NLL 1.9057, +0.1095 vs. 1.9564, +0.0588), while balanced c-BTM degrades to 1.9647.

We also experimented with learned axises instead of assigned lables. With label-free axes built from $k$-means clusters of the training prompts, \method{} keeps about 83\% of its gain ($0.090\pm0.002$ nats over three splits), although a learned dense gate reaches a lower NLL there without producing experts that work alone (Appendix~\ref{app:label-free}).

To compare with soft clustering, we give soft $k$-means the same trait space, resource information, dense share-weighted training, expert count and budget as the game, with positions frozen. The $k$-means positions are interchangeable with the game's equilibrium positions (paired difference $0.0006\pm0.0007$ nats over three splits), so the game reaches clustering-quality positions from a competitive objective. Soft $k$-means' own allocation trails the game's by 0.010 nats when one expert is sampled and by 0.002 under probability mixing, and most of that gap comes from the Dirichlet kernel (Appendix~\ref{app:soft-clustering}). Thus kernel choice and parameterization give \method{} a competitive advantage over soft clustering, and the game may possibly be a generalization of soft $k$-means to arbitrary influence kernels and trait-space geometries.

Out of distribution, we test GSM8K, which no training benchmark resembles, and seven unseen datasets of the trained kinds. Fine-tuning on this corpus costs every model much of the base model's math ability, and the game keeps the most, answering about twice as many problems as the pooled adapter (accuracy 0.41 against 0.21 over three splits). On the unseen datasets both mixtures beat pooling on full text with a dense gate slightly ahead, while on response tokens alone no mixture beats the pooled adapter (Appendix~\ref{app:ood}).

On safety benchmarks (Appendix~\ref{app:safety}) the game outperformed the pooled adapter, but the base model far outperformed all trained models because the corpus trains on unsafe responses and bare harmful prompts; retraining on safe responses returns every method to the base model's attack-success level, where the game keeps the most instruction following. Because the router is fixed, the harm can also be traced to an expert before anything is retrained: 94\% of the HarmBench prompts route to one pair, and retraining only that pair on safe responses lowers attack success from 0.51 to 0.02 while IFEval changes by $-0.011$, within noise; the same data trained into another expert changes nothing.

 \section{Limitations and Discussion}
\label{sec:limits}

The headline claim of \method{} is that the influencer's game makes specialization of experts an explicit process rather than an empirically derived one. Through experiments, we have seen that even without training a gate on experts' losses, \method{} predicted the best expert 90\% of the time as its top choice and the best expert 96\% of the time in its top-2 choices when routing data. This has many consequences for interpretability, since we may isolate how data is routed and to which experts. For safety, this means that we may target experts with specific safety data, or if the MoE experts perform poorly on a specific data point in the trait space, identify which expert is explicitly responsible and retrain it alone (Section~\ref{sec:exp:behavioral}).

Importantly, there are many levers that researchers can pull in order to shape the behavior of the MoE through the game structure, including the number of experts, the data distribution representation, and kernel choice and its parameters. Perhaps more useful are the equilibria given by the influencer's game component, which provide targets for models adapting to an unknown data distribution or possibly a changing one. If the data distribution is known, then the experts' equilibrium positions can be calculated before training time and used as stable initializations for expert specialization.   
There are also limitations. \method{} relies on symmetric and log-concave kernels, which implies that experts cannot have multiple central points of expertise and that their influence cannot decay asymmetrically. While the first can be addressed by a larger number of experts, there is a real cost to training more adapters. The second limitation is more restrictive because the theory fundamentally relies on symmetric decay of influence, but the need for asymmetric kernels remains uncertain. The game also routes per prompt, since it needs one fixed trait vector per data point; over conventional sparse-upcycled experts its shares come within 0.006 nats of a learned dense router, but no token-level router we trained matched the pooled adapter (Appendix~\ref{app:upcycle}).

Perhaps the most promising future direction is online adaptation to new data streams. Because there is no trained gate, there is no learned router to collapse. The game defines a position payoff for any observed resource distribution, so an updated prompt distribution implies new position incentives and a new candidate equilibrium. In the present fixed-distribution experiments, the position dynamics already correct an initial position mismatch (Section~\ref{sec:kernel-params}), although this does not test distribution shift. A direct test would change the prompt distribution over time, compare predicted and observed position movement, and continue adapter training under the revised allocations.
 \section{Conclusion}
\label{sec:conclusion}
\method{} treats expert coordination as a location game over a trait space of a data distribution. Its equilibrium supplies a starting position before any adapter is trained, and its allocation is a falsifiable prediction about held-out expert quality rather than a fitted gate. The components of \method{} are not limited to the language model setting or by choices such as the encoder and data type, and may contribute to other areas of work. The open question is how this method will apply to online settings where the data distribution is not fixed. 
 
\section{AI Use Disclosure}
\label{sec:ai-use}
Generative AI tools, including OpenAI Codex, assisted with literature discovery and citation checking, manuscript organization and language editing, and software implementation and debugging. The authors reviewed all retained prose and code, checked citation-dependent claims against primary sources, tested AI-assisted implementation, and verified reported numerical claims against saved experimental artifacts. AI-generated output was not treated as empirical evidence or credited with authorship; the authors take responsibility for the paper and all associated artifacts.

\section{Ethics Statement}
\label{sec:ethics}
This work develops a routing and training-allocation method and conducts no human-subject experiments or new collection of personal data; the experiments use existing public benchmarks whose provenance is documented in Appendix~\ref{app:data-provenance}. The method may make expert behavior easier to audit, but it does not guarantee safety or fairness, and specialization can concentrate failures as well as useful capabilities. Appendix~\ref{app:safety} reports the observed safety limitations, including that fine-tuning substantially degraded alignment relative to the base model.

\section{Reproducibility Statement}
\label{sec:reproducibility}
The game assumptions and stability analysis are stated in Section~\ref{sec:game} and proved in Appendix~\ref{app:derivations}; the data sources, retained counts, shared splits, model and adapter settings, routing procedure, and evaluation estimators are documented in Section~\ref{sec:design} and Appendices~\ref{app:data-provenance}--\ref{app:routing-ablation}. Appendices~\ref{app:model-ablations}, \ref{app:baselines}, and \ref{app:safety} specify the ablation, baseline, and behavioral protocols and their qualifications. The accompanying code uses versioned YAML configurations, records the resolved configuration and run history, and exposes a single pipeline entry point; Appendix~\ref{app:repro} gives the command and identifies the result artifacts underlying the reported analyses. Access to the code is available here: \url{https://anonymous.4open.science/r/infl_ens-2516/README.md}.

\bibliographystyle{iclr2027_conference}

\newpage
\appendix

\section{Data provenance}
\label{app:data-provenance}
Table~\ref{tab:resource-data} lists the seven public datasets that make up the resource population, one per trait axis, with the retained sample count and the source annotation used to estimate each direction. The signals label the prompt, not our models: the over-refusal signal, for instance, distinguishes OR-Bench's benign-but-seemingly-toxic prompts from its toxic controls, not whether one of our models refused.

\begin{table}[!h]
\centering
\small
\setlength{\tabcolsep}{3pt}
\caption{Data used to construct the resource landscape. Counts are the records retained before the shared split. ``Signal'' is the annotation used to learn the corresponding trait direction.}
\label{tab:resource-data}
\begin{tabular}{@{}>{\raggedright\arraybackslash}p{0.15\linewidth}
                    >{\raggedright\arraybackslash}p{0.29\linewidth}
                    >{\raggedright\arraybackslash}p{0.43\linewidth}r@{}}
\toprule
Trait & Source & Signal & $n$ \\
\midrule
Harm & BeaverTails-30k \citep{ji2023beavertails}
  & response unsafe in any harm category & 5,000 \\
Hallucination & HaluEval QA and dialogue \citep{li2023halueval}
  & faithful versus hallucinated response & 5,000 \\
Jailbreak & JBB-Behaviors \citep{chao2024jailbreakbench}
  & harmful versus benign goal & 200 \\
Privacy & AI4Privacy PII-Masking-200k \citep{ai4privacy2023}
  & character fraction covered by PII spans & 5,000 \\
Over-refusal & OR-Bench 80k and toxic \citep{cui2024orbench}
  & seemingly toxic benign prompt versus toxic control & 3,155 \\
Injection & Prompt Injection Threat Matrix \citep{jajoo2026threatmatrix}
  & binary malicious-injection label & 5,000 \\
Policy violation & Do-Not-Answer \citep{wang2023donotanswer} with
  Alpaca-Cleaned negatives \citep{yahma2023alpacacleaned}
  & refusal-worthy versus benign instruction & 5,000 \\
\bottomrule
\end{tabular}
\end{table}

\paragraph{What the adapters are trained to produce.}
The fine-tuning loss covers every non-padding token of the chat text, prompt included, and the corpus keeps both labels of its three response-bearing sources. Table~\ref{tab:corpus-composition} lists the resulting targets in the seed-0 training partition. About 10\% of the 19,848 training records carry an unsafe response as their target and 9\% a hallucinated answer, and 47\% are prompts with no response, on which the model is trained to reproduce the prompt. The test partition has the same composition: 2,671 of its 5,671 records come from the four prompt-only sources and contribute prompt tokens alone, so every NLL in this paper scores full chat text. Every arm, the pooled adapter included, trains and is scored on this corpus. Appendix~\ref{app:safety} traces the behavioral degradation after fine-tuning to these targets and retrains without them.

\begin{table}[!h]
\centering
\small
\caption{Training targets by source, seed-0 training partition (19,848 records).}
\label{tab:corpus-composition}
\begin{tabular}{@{}lrl@{}}
\toprule
Source & Train & Training target \\
\midrule
BeaverTails & 3,500 & 1,999 unsafe responses, 1,501 safe responses \\
HaluEval & 3,500 & 1,747 hallucinated answers, 1,753 correct answers \\
AI4Privacy & 3,500 & privacy-masking responses \\
Prompt Injection Threat Matrix & 3,500 & the prompt itself (2,588 injection attempts) \\
Do-Not-Answer with Alpaca-Cleaned & 3,500 & the prompt itself \\
OR-Bench & 2,208 & the prompt itself \\
JBB-Behaviors & 140 & the prompt itself (75 harmful behaviors) \\
\bottomrule
\end{tabular}
\end{table}

\section{Game-theoretic derivations}
\label{app:derivations}

The stability condition of Section~\ref{sec:game:stability} and the pairing result used in Section~\ref{sec:exp:setup} are due to \citet{lovett2026influencer}; this appendix repeats their derivations in the notation of this paper.

\paragraph{Regularity and symmetric stationarity.}
Let all $\nspec\geq2$ players use the same positive kernel $\kernel(\cdot,b;\reach)\in C^2$ on an open neighborhood of their strategy domain, and suppose differentiation can pass through the expectation over $\resource$. The formulas below are Euclidean; for simplex-valued positions, all gradients, Hessians, and definiteness statements are restricted to the simplex tangent space.

Write $s_i(b)=\nabla_{x_i}\log\kernel(x_i,b;\reach)$ and $H_i(b)=\nabla_{x_i}^2\log\kernel(x_i,b;\reach)$. Differentiating the normalized share gives
\begin{equation}
\nabla_{x_i}\share{i}=\share{i}(1-\share{i})s_i,\qquad
\nabla_{x_j}\share{i}=-\share{i}\share{j}s_j\quad(j\neq i),
\label{eq:share-derivatives}
\end{equation}
and hence the positional gradient field is
\begin{equation}
F_i(\allpos):=\nabla_{x_i}\util{i}(\allpos)
 =\expectB{\share{i}(1-\share{i})s_i}.
\label{eq:game-gradient}
\end{equation}
At a symmetric profile $x_1=\cdots=x_\nspec=x^\circ$, every share is $1/\nspec$, so
\begin{equation}
\nabla_{x_i}u_i
 =\frac{\nspec-1}{\nspec^2}\expectB{s(b)}.
\label{eq:symmetric-gradient}
\end{equation}
Thus an interior symmetric profile is stationary exactly when $\expectB{s(b)}=0$. Equivalently, it is a stationary point of
\begin{equation}
\Phi(x)=\expectB{\log\kernel(x,b;\reach)}.
\label{eq:expected-log-influence}
\end{equation}
If the strategy domain is compact, $\Phi$ is continuous, and its maximizer is interior, such a point exists; if $\Phi$ is strictly concave, it is unique. On a boundary, Equation~\plaineqref{eq:symmetric-gradient} is replaced by the projected first-order condition. These conditions do not certify a global Nash equilibrium or a separated solution, which require best-response or numerical checks.

\paragraph{Jacobian and player-separating modes.}
At $x^\circ$, let
\begin{equation}
\bar H=\expectB{H(b)},\qquad
\scatter=\expectB{s(b)s(b)^\top},\qquad
\curv=-\bar H .
\label{eq:theory-moments}
\end{equation}
At a stationary point $\scatter$ is the score covariance. Differentiating Equation~\plaineqref{eq:game-gradient} once more and substituting $\share{i}=1/\nspec$ gives identical diagonal and off-diagonal blocks,
\begin{align}
A:=\frac{\partial F_i}{\partial x_i^\top}
 &=\frac{\nspec-1}{\nspec^2}\bar H
   +\frac{(\nspec-1)(\nspec-2)}{\nspec^3}\scatter,\notag\\
\Delta:=\frac{\partial F_i}{\partial x_j^\top}
 &=-\frac{\nspec-2}{\nspec^3}\scatter,\qquad j\neq i.
\label{eq:jacobian-blocks}
\end{align}
Consequently, the full $\nspec\ndim\times\nspec\ndim$ Jacobian has $A$ on its diagonal and $\Delta$ off its diagonal. Equivalently, \[ J=I_{\nspec}\otimes A+ (\mathbf{1}\mathbf{1}^\top-I_{\nspec})\otimes\Delta . \] The player-index space splits into the collective direction $\operatorname{span}\{\mathbf{1}\}$ and its orthogonal, sum-zero subspace. The corresponding $\ndim\times\ndim$ blocks are
\begin{align}
J_+&=A+(\nspec-1)\Delta
   =-\frac{\nspec-1}{\nspec^2}\curv,\notag\\
J_-&=A-\Delta
   =\frac{\nspec-1}{\nspec^2}
     \left[-\curv+\frac{\nspec-2}{\nspec-1}\scatter\right],
\label{eq:jacobian-modes}
\end{align}
where $J_+$ appears once and $J_-$ appears with multiplicity $\nspec-1$. To verify the decomposition, choose an orthogonal matrix $P$ whose first row is $\mathbf{1}^\top/\sqrt{\nspec}$. Then $P\mathbf{1}\mathbf{1}^\top P^\top =\operatorname{diag}(\nspec,0,\ldots,0)$, and conjugating $J$ by $P\otimes I_{\ndim}$ yields $\operatorname{diag}(J_+,J_-,\ldots,J_-)$.

Assume $\curv\succ0$ on the relevant space. The collective block is then negative definite. Congruence of the separating block with \[ -I+\frac{\nspec-2}{\nspec-1} \curv^{-1/2}\scatter\curv^{-1/2} \] shows that $J_-\prec0$ exactly when the depth in Equation~\plaineqref{eq:depth} is below one. Thus $\depth<1$ gives local asymptotic stability under $\dot x_i=F_i(\allpos)$, $\depth>1$ gives a separating direction with positive linear growth, and $\depth=1$ is the zero-eigenvalue boundary. This proves the stability claim in Section~\ref{sec:game:stability}. In one dimension the same condition is
\begin{equation}
-\expectB{H(b)}>
\frac{\nspec-2}{\nspec-1}\expectB{s(b)^2}.
\label{eq:one-dimensional-stability}
\end{equation}

\paragraph{Two-player and Gaussian consequences.}
When $\nspec=2$, the score-scatter terms and $\Delta$ vanish: $J_+=J_-=-\curv/4$. Hence any interior symmetric stationary point with $\curv\succ0$ is locally asymptotically stable.

For the elliptical Gaussian kernel, let $Q\succ0$ and \[ \log\kernel(x,b;\reach) =-\frac{(b-x)^\top Q(b-x)}{2\reach^2}+\mathrm{const}. \] Then $s=Q(b-x)/\reach^2$ and $H=-Q/\reach^2$. Equation~\plaineqref{eq:symmetric-gradient} gives $x^\circ=\expectB{b}$ for an unconstrained interior solution. Writing $\Sigma_B=\operatorname{Cov}_{B}(b)$, \[ \curv=\frac{Q}{\reach^2},\qquad \scatter=\frac{Q\Sigma_BQ}{\reach^4}, \] so the stability boundary has the closed form
\begin{equation}
\reachstar=
\left[
\frac{\nspec-2}{\nspec-1}
\lambda_{\max}\!\left(Q^{1/2}\Sigma_BQ^{1/2}\right)
\right]^{1/2}.
\label{eq:gaussian-stability-threshold}
\end{equation}
The symmetric Gaussian profile is stable for $\reach>\reachstar$ and separating-unstable for $\reach<\reachstar$. With $Q=I$ this becomes $\reachstar=\sqrt{\frac{\nspec-2}{\nspec-1} \lambda_{\max}(\Sigma_B)}$; in one dimension it reduces to $\reachstar=\sqrt{\frac{\nspec-2}{\nspec-1}}\,\sigma_B$.

\paragraph{Exact reduction for co-located groups.}
Suppose each effective player $p$ is represented by $r$ equal-size clones with the same position and kernel value $f_p(b)$. Each clone receives $f_p(b)/(r\sum_q f_q(b))$, and summing its group gives
\begin{equation}
\sum_{a=1}^{r}G_{p,a}(b)
=\frac{r f_p(b)}{r\sum_q f_q(b)}
=\frac{f_p(b)}{\sum_q f_q(b)}.
\label{eq:clone-reduction}
\end{equation}
Therefore equal-size groups that remain co-located have exactly the shares and summed utilities of the reduced game with one player per group, so taking $r=2$ duplicates each solved location without perturbing the effective allocation. With the two-player stability result above, this motivates initializing the experts in pairs. The reduction is an identity on the co-located subspace; two-player stability does not prove that two unconstrained members embedded in a larger game would remain together. The reduction also fails when group sizes differ. Appendix~\ref{app:position-update} shows why the implemented update preserves equal co-location and distinguishes the clone-level gradient from a jointly deviating reduced-game player.

\section{General-kernel game-gradient updates}
\label{app:position-update}

The position update uses the payoff in Equation~\plaineqref{eq:allocation}, not language-model loss. Let $s_i(b)=\nabla_{x_i}\log\kernel(x_i,b;\reach)$. The exact population vector field derived in Appendix~\ref{app:derivations} is
\begin{equation}
 \nabla_{x_i}u_i(\allpos)
 =\expectB{\share{i}(1-\share{i})s_i(b)}.
 \label{eq:population-position-gradient}
\end{equation}
At round $t$, a batch $b_1,\ldots,b_M$ replaces the resource expectation, giving the stochastic field
\begin{equation}
 \widehat g_i^{(t)}=
 \frac{1}{M}\sum_{m=1}^{M}
 \share{i}(\allpos^{(t)},b_m)
 \bigl[1-\share{i}(\allpos^{(t)},b_m)\bigr]s_i(b_m).
 \label{eq:batch-position-gradient}
\end{equation}
For batches sampled from $\resource$, this is an unbiased estimator of Equation~\plaineqref{eq:population-position-gradient}.

\paragraph{General update.}
The method applies the simultaneous projected step
\begin{equation}
 \allpos^{(t+1)}=
 \Pi_{\traitspace}\!\left(
 \allpos^{(t)}+\bar\eta_t
 P_{T\traitspace}\widehat g^{(t)}
 \right).
 \label{eq:projected-game-step}
\end{equation}
Each influence family supplies its own score $s_i(b)$ using the closed forms in Appendix~\ref{app:kernels}; no centroid assumption is required. $P_{T\traitspace}$ removes components normal to the domain (notably the all-ones direction on the simplex), and $\Pi_{\traitspace}$ restores feasibility. A single population-wide effective step size $\bar\eta_t$ is reduced when necessary so that the largest player movement respects the step cap. Before the final feasibility projection, this scalar preserves the joint game-gradient direction and the relative magnitudes of the players' updates. The kernel-family experiments use this rule with the observed round batch.

Because the batch defines the empirical resource measure in Equation~\plaineqref{eq:batch-position-gradient}, a changing prompt distribution changes the vector field in a predictable way. Repeated projected steps can therefore track a moving resource landscape without learning a new router objective.

\paragraph{Gaussian centroid as a special case.}
For an isotropic Gaussian kernel, $s_i(b_m)=(b_m-x_i)/\reach^2$. Define
\begin{equation}
 w_{im}=\share{i}(\allpos,b_m)\bigl[1-\share{i}(\allpos,b_m)\bigr],
 \quad W_i=\sum_{m=1}^{M}w_{im},
 \quad \widehat x_i=\frac{\sum_m w_{im}b_m}{W_i}.
 \label{eq:matched-centroid}
\end{equation}
Then
\begin{equation}
 \widehat g_i=\frac{W_i}{M\reach^2}(\widehat x_i-x_i).
 \label{eq:centroid-gradient-equivalence}
\end{equation}
The reported principal Gaussian arms evaluate this direction through the blended centroid step
\begin{equation}
 x_i^{(t+1)}=(1-\beta)x_i^{(t)}+\beta\widehat x_i
 =x_i^{(t)}+\eta_i^{(t)}\widehat g_i^{(t)},
 \qquad
 \eta_i^{(t)}=\frac{\beta M\reach^2}{W_i}>0,
 \label{eq:implemented-position-step}
\end{equation}
with $\beta=0.5$. This specialization is exactly parallel to the empirical Gaussian utility gradient and has the same stationary points whenever $W_i>0$. Its step size is an agent- and batch-adaptive positive rescaling rather than the population-wide learning rate in Equation~\plaineqref{eq:projected-game-step}, so the two implementations do not have identical step magnitudes.

The direct general-kernel rule evaluates Equation~\plaineqref{eq:batch-position-gradient} for every clone and every prompt in the round batch. Likewise, the dense Gaussian specialization evaluates $w_{im}$ on the whole batch. Neither rule truncates its position mass to the top-$k$ training assignments or renormalizes it across retained pairs.

The reported hard-routing Gaussian arm uses the corresponding stochastic centroid estimator. If $Z_{im}$ indicates that prompt $m$ was sampled for clone $i$, then
\begin{equation}
 \mathbb{E}\!\left[Z_{im}(1-\share{i})\mid b_m\right]
 =\share{i}(1-\share{i}).
 \label{eq:hard-position-mass}
\end{equation}
The selected prompts are therefore weighted by $(1-\share{i})$: their unnormalized centroid mass estimates the same theoretical coefficient, but the finite-batch ratio is noisier than evaluating the full batch. Updated positions affect routing beginning with the next round.

\paragraph{Why pairs remain co-located.}
For two equal clones $a$ and $a'$ at the same position, $\share{a}(b_m)=\share{a'}(b_m)$ for every prompt. Their rows of $w_{im}$, their centroid targets, and therefore their independently computed updates are identical. Exact co-location is an invariant of the dense theory-matched dynamics; the implementation does not average a pair and write one forced position back to both members.

\begin{figure}[t]
  \centering
  \includegraphics[width=\linewidth]{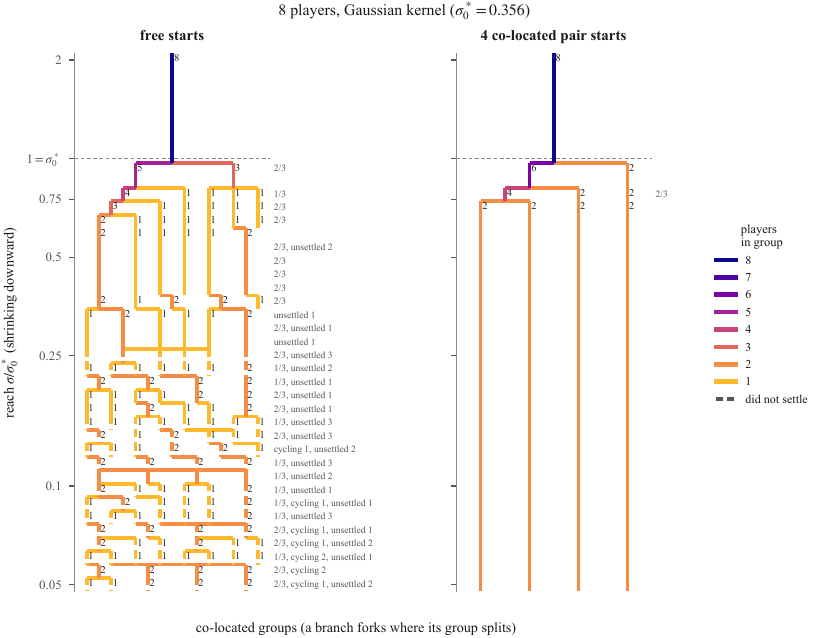}
  \caption{Co-located groups of eight players in the Gaussian position game against the reach, with no adapter training. Each branch is a group of co-located players, colored and labeled by its member count, and forks where the group splits; the reach shrinks downward. Left, every player starts at a random prompt; right, the players start as four co-located pairs. Both trees use the same 41 reaches and three seeds per reach, and each row draws the most common partition. Dashed branches did not settle within 8,000 steps; the right margin gives seed agreement and the number of seeds that cycled or did not settle.}
  \label{fig:group-tree}
\end{figure}

If $G_p$ denotes the summed share of a two-clone pair, then each member has $G_i=G_p/2$ and follows the coefficient $G_i(1-G_i)$. A hypothetical player that moved the two clones jointly to maximize their summed utility would instead have coefficient $G_p(1-G_p)$. Equation~\plaineqref{eq:clone-reduction} therefore gives an exact reduction of allocations and summed utilities on the co-located subspace, but not an identity between these two gradient fields. The implemented update matches each clone's own utility gradient in the fourteen-player game, while routing and adapter training operate on seven aggregated pairs.
\paragraph{Group structure of the position game.}
Figure~\ref{fig:group-tree} solves the position game alone, with no adapter training, for eight individual players under an isotropic Gaussian kernel on the empirical measure of all 28,355 prompts, for which $\reachstar=0.356$. Each of 41 reaches from $2\reachstar$ to $0.05\reachstar$ is solved independently from the same seeded start with the blended centroid step of Equation~\plaineqref{eq:implemented-position-step}, for up to 8,000 steps and without perturbation, on three seeds. Players within $\max(0.075,\,0.35\reach)$ of one another are counted as one group. From free starts, each player at a random prompt, the eight players stay together down to $\reachstar$, separate into five and three just below it and fragment into pairs and single players by $0.6\reachstar$; below about $0.3\reachstar$ many solves do not settle within the budget and the seeds disagree. From four co-located pairs, the players pass through $6+2$ and $4+2+2$ and settle into four pairs from about $0.7\reachstar$ to $0.05\reachstar$; every one of these solves converges and no pair separates, as the invariance above requires.

\section{Trait-space geometry}
\label{app:trait-geometry}
\begin{figure}[t]
  \centering
  \includegraphics[width=\linewidth]{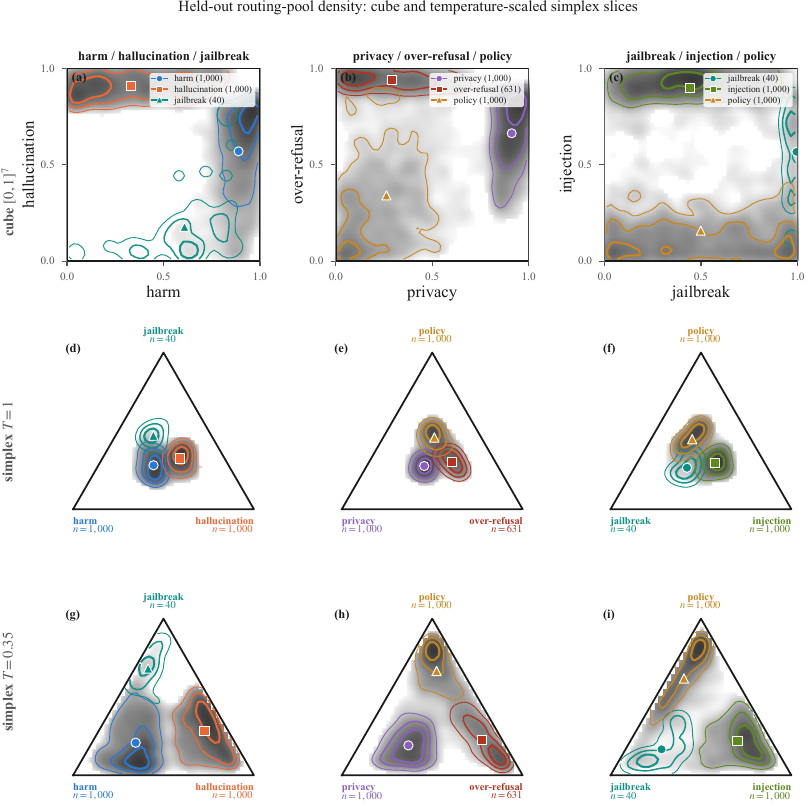}
\caption{The trait projection controls how much geometric separation the kernel can exploit. Columns show three benchmark groupings among 5,671 held-out routing prompts: two-coordinate cube projections (top), followed by conditional three-component simplex views at $T=1$ (middle) and $T=0.35$ (bottom). Grey is the combined prompt density; colored contours and markers denote benchmark origin, with each class normalized to its own density mode and therefore not encoding prevalence. Each triangle renormalizes only its three named components; the other four coordinates are not assumed to be zero. The $T=1$ transform compresses prompts near the simplex center, whereas $T=0.35$ spreads them toward distinct vertices. These are resource densities, not expert territories, and demonstrate why denser geometry requires a more peaked influence kernel to obtain selective routing.}
  \label{fig:trait-slices}
\end{figure}

The cube retains seven independently calibrated coordinate levels, whereas the simplex encodes their relative composition. For calibrated scores $c\in[0,1]^7$, the latter uses
\begin{equation}
 b_j(T)=\frac{\exp(c_j/T)}{\sum_{\ell=1}^{7}\exp(c_\ell/T)},
 \qquad T>0.
 \label{eq:trait-temperature}
\end{equation}
Figure~\ref{fig:trait-slices} shows the held-out resource density in these spaces. Each ternary view conditionally renormalizes its three named components; the other four components of the full seven-axis representation are not assumed to be zero. At $T=1$ the bounded logits confine each displayed ternary component to approximately $[0.155,0.576]$, crowding the prompt cloud into the interior. At $T=0.35$ that possible range expands to approximately $[0.028,0.897]$, and the measured support spreads toward the corners.

This geometry affects how peaked an influence kernel must be to distinguish competitors. For the reported Gaussian arms with identity shape matrix, the allocation log-odds obey
\begin{equation}
 \log\frac{G_i(b)}{G_j(b)}
 =\frac{\lVert b-x_j\rVert^2-\lVert b-x_i\rVert^2}{2\sigma^2}.
 \label{eq:trait-kernel-odds}
\end{equation}
When resources are concentrated where rival positions have similar affinity, a broad kernel makes these odds small and spreads weight across experts. Narrowing effective reach, increasing kernel tempering, or using a less compressive trait projection can produce a more discriminative allocation. This can lower expected single-expert NLL when diffuse routing gives weight to weaker experts, as the sharpness-matched $T=1$ comparison and the $T=0.35$ end-to-end arms in Figure~\ref{fig:init-kernels} suggest. A compact cloud already close to one expert can route sharply, and extra peaking need not improve probability mixing or preserve adequate training mass for every expert.

With a fixed trait encoder and expert family, reach $\sigma$ or a tempering exponent can be validated as router controls while monitoring allocation entropy, expert loads, and the intended inference NLL. Trait temperature $T$ is a separate geometry lever, but changing it also changes the position game; positions must be re-solved or recalibrated, and the resulting router tested against the same frozen experts. The reported $T=0.35$ arms instead retrained experts under the new geometry, so they do not establish a free post-training gain.

An end-to-end simplex experiment shows higher router--oracle agreement (about 0.87 versus 0.71) and mixture capture (91.5--92.6\% versus 82.0--84.9\%) than its cube counterpart. A bidirectional frozen-expert cross-routing control qualifies that comparison. Simplex-trained experts score 1.9383 NLL under their own geometry versus 2.0739 under cube routing, while cube-trained experts score 1.9465 under cube versus 2.0244 under simplex, so each family prefers the geometry in which it trained. Absolute reach and a density-estimation routine also differ between the two arms, and one reverse solve did not converge.

Appendix~\ref{app:kernels} gives the kernel closed forms and the distinct ways their parameters control reach and sharpness.

\section{An inductive trait space}
\label{app:inductive}

Every other trait space in this paper is fitted on all 28{,}355 records (Section~\ref{sec:exp:setup}): the axis directions on all benchmark labels, and the quantile maps and resource KDE on all prompts, so the test prompts help shape the space that routes them.
We refit all three on the training partition only (\texttt{trait\_space.fit\_partition: train}) and rerun the seven-pair $T=0.35$ Dirichlet arm with nothing else changed, on three data splits; reach is held at its transductive value, and the start is re-solved inside the train-only space so no position carries test information.
The train-only axes differ noticeably from the originals ($|\cos|$ 0.86--0.97) while the resource distribution moves little (total variation 2.2--2.4\%).
Table~\ref{tab:inductive} shows the result: the share-weighted single expert loses $-0.0091\pm0.0027$ nats, with the same sign on every split, about 8\% of the gain; the mixture loses $-0.0042$ and top-1 agreement -1.6 points, while the hindsight-best expert does not move ($0.0000\pm0.0009$).
The transductive fit therefore inflates routing, not expert quality, and routed levels elsewhere in the paper should be read as about 0.007--0.012 nats optimistic; comparisons between arms that share one trait space give both sides the same advantage.

\begin{table}[h]
  \centering
  \caption{Trait space fitted on all records (all) or on the training partition only (train), seven-pair $T=0.35$ Dirichlet arm, test partition. Gains in nats (higher is better); top-1 in percent.}
  \label{tab:inductive}
  \small
  \setlength{\tabcolsep}{4pt}
  \begin{tabular}{lrrrrrrrr}
    \toprule
    & \multicolumn{2}{c}{Single expert} & \multicolumn{2}{c}{Mixture} & \multicolumn{2}{c}{Oracle} & \multicolumn{2}{c}{Top-1} \\
    \cmidrule(lr){2-3} \cmidrule(lr){4-5} \cmidrule(lr){6-7} \cmidrule(lr){8-9}
    Split & all & train & all & train & all & train & all & train \\
    \midrule
    seed 0 & 0.1168 & 0.1048 & 0.1286 & 0.1227 & 0.1427 & 0.1423 & 90.1 & 86.9 \\
    dseed1 & 0.1165 & 0.1080 & 0.1306 & 0.1270 & 0.1471 & 0.1482 & 88.7 & 87.6 \\
    dseed2 & 0.1217 & 0.1150 & 0.1332 & 0.1302 & 0.1465 & 0.1458 & 89.4 & 89.0 \\
    \bottomrule
  \end{tabular}
\end{table}

\section{Label-free trait axes}
\label{app:label-free}

The seven axes of Section~\ref{sec:exp:setup} are fitted on benchmark annotations. Here we replace them with seven directions of the same prompt embeddings that use no labels and are fitted on the training partition only. Everything downstream is unchanged: the quantile maps, the $T=0.35$ simplex, the resource distribution, one expert pair per axis, and a reach re-solved to the principal arm's instability depth. Each arm is retrained on three data splits and scored against its own split's pooled adapter.

\paragraph{Cluster-derived axes.}
The game places experts on the modes of the resource distribution, so our main label-free space is built to find modes. The training prompts' embeddings are reduced to their top 50 principal components and grouped by $k$-means with $K=7$ clusters and 10 restarts. One shrinkage Fisher direction then separates each cluster from the remaining prompts, orthogonalized against the directions already found. This is the labelled construction of Section~\ref{sec:exp:setup} with benchmark identity replaced by cluster identity, and we refer to the result as pseudo-Fisher axes. The clusters recover most of the benchmark structure without seeing it: each is mostly one benchmark, with mean purity 0.82.

\paragraph{PCA and ICA axes.}
Two simpler spaces serve as comparisons: the top seven principal components of the training prompts' embeddings, and the seven most non-Gaussian of 20 whitened independent components.

\begin{figure}[h]
  \centering
  \includegraphics[width=\linewidth]{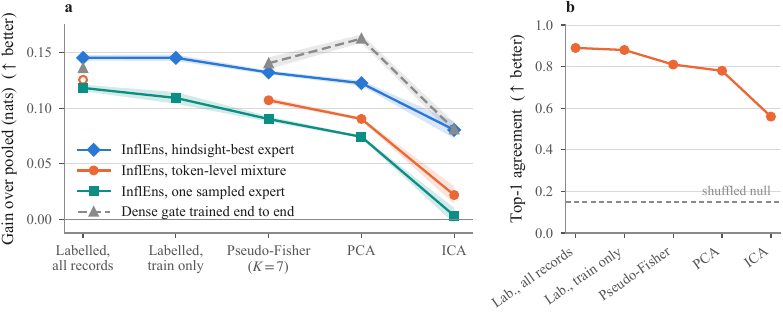}
  \caption{The game on labelled and label-free trait axes, seven pairs, $T=0.35$. \textbf{a}~Gain over each split's pooled adapter; markers are means over three data splits and bands one sample standard deviation. One sampled expert is the share-weighted loss of sampling one pair, and the hindsight-best expert is the oracle. The dense gate is a dense softmax gate trained end to end on the same axes with one expert per axis; it was not trained on the train-only labelled axes. The hollow marker is the token-level mixture on the labelled axes, which exists for the seed-0 split only. The dense gate's lead over the token-level mixture is $0.034\pm0.006$ nats on pseudo-Fisher axes, $0.072\pm0.004$ on PCA and $0.059\pm0.005$ on ICA, and 0.012 on the labelled axes of the seed-0 split. \textbf{b}~Fraction of test prompts whose best expert receives the largest share, against a shuffled null of 0.15.}
  \label{fig:label-free}
\end{figure}

\begin{figure}[t]
  \centering
  \includegraphics[width=0.9\linewidth]{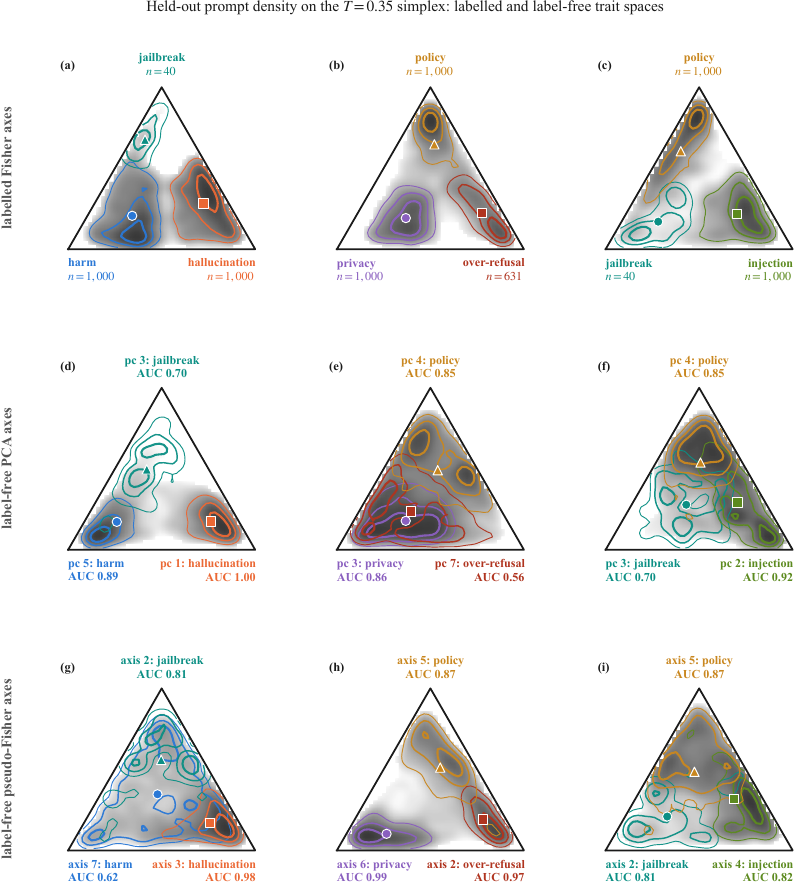}
  \caption{Seed-0 held-out prompts on the three benchmark triplets of Figure~\ref{fig:trait-simplex}, as conditional views of three simplex components at $T=0.35$. Grey is the smoothed density of the triplet's prompts; colored contours and centroid markers give the benchmark of origin. Rows are labelled Fisher axes (a--c), label-free PCA axes (d--f) and label-free pseudo-Fisher axes (g--i), the last two fitted on the training partition only. In the label-free rows each corner is the axis that best separates that benchmark, with its AUC. On PCA axes privacy and over-refusal share one region (e) and injection overlaps jailbreak (f); on pseudo-Fisher axes they return to their own corners.}
  \label{fig:label-free-slices}
\end{figure}

\paragraph{Routing gain.}
Figure~\ref{fig:label-free} gives the result. On pseudo-Fisher axes the game keeps about 83\% of its train-only labelled gain and identifies each prompt's best expert 81\% of the time. PCA keeps about 70\%, and the pseudo-Fisher game improves on it on every split and estimator (paired, $0.016\pm0.002$ nats expected and $0.017\pm0.002$ token level). PCA models the data as one ellipsoid and returns directions of largest overall variance, which run through several clusters at once, so the prompts spread over the interior of the simplex and the game has fewer distinct modes to compete over (Figure~\ref{fig:label-free-slices}). ICA gives almost no routing gain. The axes therefore need not be hand-assigned, but they must carry task-relevant, multimodal variation. Ten clusters are worse than seven for the game on the seed-0 split (expected gain 0.0581), where one pair receives 0.2\% of the load and the extra axes force a three times smaller reach. Six pairs on the seven pseudo-Fisher axes change the seed-0 expected gain from 0.0908 to 0.0867. Over three training seeds on the seed-0 split, the pseudo-Fisher game's token-level mixture scores $1.9070\pm0.0011$.

\paragraph{What the label-free axes encode.}
Each PCA axis tracks one benchmark, with the same assignment on all three splits. Benchmark-versus-rest AUC is 0.95 for over-refusal on the first component, 0.92 for prompt injection, 0.89 for privacy, 0.84 for policy violation, 0.84 for harm, and 0.78 and 0.71 for hallucination on the last two; only jailbreak has no axis of its own. The PCA subspace is nonetheless nearly orthogonal to the labelled Fisher subspace (principal angles 76--90$^\circ$), so different directions of the 4096-dimensional embedding carry overlapping task information. On pseudo-Fisher axes privacy, over-refusal and policy violation separate with AUC 0.99, 0.97 and 0.87, and the one impure cluster is harm (purity 0.75). Expert purity follows the axes: six of seven labelled experts win at least 85\% of their prompts from one benchmark (Table~\ref{tab:expert-names}), against two of seven on PCA axes and none on ICA axes.

\paragraph{Learned gates on the same axes.}
A dense softmax gate trained end to end on each label-free space, at router learning rate $10^{-2}$ and a $1\times16$ micro-batch, reaches a lower NLL than the game (dashed line in Figure~\ref{fig:label-free}a). Its lead is $0.072\pm0.004$ nats on PCA axes and halves to $0.034\pm0.006$ on pseudo-Fisher axes, where the gate itself does $0.022\pm0.006$ worse than on PCA. Over three training seeds on the seed-0 split the pseudo-Fisher dense gate scores $1.8740\pm0.0054$, a paired gap to the game of $0.033\pm0.005$, so the game is about five times more seed-stable. As on labelled axes (Appendix~\ref{app:gates}), the gate's components do not specialize: on PCA axes its hindsight-best component (2.119) is worse than its blended network, and every component alone is worse than the pooled adapter. The other gates trained end to end on the pseudo-Fisher axes on the seed-0 split score 1.8775 (Gaussian), 1.8966 (DeepSeek-V3), 1.9167 (Mixtral top-2) and 1.9753 (Switch top-1), against 1.8692 for the dense gate and 1.9074 for the game's token-level mixture, and their components also work only when blended. c-BTM loses its whole gain on PCA axes (2.0196 against the pooled 2.0154), so the game's lead over it widens from 0.058 nats on labelled axes to 0.079.

\paragraph{Gates fitted on the game's label-free experts.}
Table~\ref{tab:label-free-gates} repeats the post hoc comparison of Appendix~\ref{app:gates} on the label-free spaces: the published gates are fitted on validation losses over the game's frozen experts and scored on the test pool. The distance gates again route the game's experts slightly better than the game's own shares, by 0.005 nats on pseudo-Fisher axes and 0.008 on PCA as a sequence mixture. Mixed per token on the pseudo-Fisher experts, the X-MoE and Gaussian gates score 1.8949 and 1.8950 against 1.9071 for the game's shares. Better routing over the game's experts therefore closes 0.012 of the dense gate's 0.038 seed-0 lead on these axes, and the remainder lies in what the jointly trained network learns.

\begin{table}[h]
  \centering
  \caption{Gates fitted post hoc on the game's frozen label-free experts, seed-0 split, test NLL (lower is better; pooled adapter 2.0154). Fitted on the 2,836 validation records, means over three gate seeds. Top-1 is agreement with the hindsight-best expert.}
  \label{tab:label-free-gates}
  \small
  \setlength{\tabcolsep}{4pt}
  \begin{tabular}{@{}lcccccc@{}}
    \toprule
    & \multicolumn{3}{c}{Pseudo-Fisher experts} & \multicolumn{3}{c}{PCA experts} \\
    \cmidrule(lr){2-4}\cmidrule(lr){5-7}
    Gate & Seq.\ mixture & Argmax & Top-1 & Seq.\ mixture & Argmax & Top-1 \\
    \midrule
    \oursrow \method{} shares (nothing fitted) & 1.9026 & 1.9220 & 0.823 & 1.9189 & 1.9391 & 0.782 \\
    Gaussian & 1.8975 & 1.9146 & 0.844 & 1.9116 & 1.9375 & 0.785 \\
    X-MoE cosine & 1.8975 & 1.9150 & 0.844 & 1.9113 & 1.9372 & 0.785 \\
    Dense softmax & 1.8997 & 1.9198 & 0.827 & 1.9132 & 1.9392 & 0.781 \\
    DeepSeek-V3 & 1.9022 & 1.9260 & 0.817 & 1.9183 & 1.9405 & 0.779 \\
    Switch top-1 & 1.9142 & 1.9142 & 0.846 & 1.9365 & 1.9365 & 0.784 \\
    Mixtral top-2 & 2.0835 & 2.1254 & 0.375 & 2.1172 & 2.1406 & 0.421 \\
    Oracle (best expert per prompt) & 1.8853 & 1.8853 & 1.000 & 1.8952 & 1.8952 & 1.000 \\
    \midrule
    Dense gate trained end to end, blended & \multicolumn{3}{c}{1.8692} & \multicolumn{3}{c}{1.8540} \\
    \bottomrule
  \end{tabular}
\end{table}

\section{Influence kernels and design rationale}
\label{app:kernels}
All kernel families used by the game specify a positive influence $f_\kappa(x,b)$; Equation~\plaineqref{eq:allocation} normalizes their values across players. The solver uses the position score $s=\nabla_x\log f_\kappa$ and log-Hessian $H=\nabla_x^2\log f_\kappa$ to evaluate the game gradient and the local instability criterion. Below $\sigma>0$ is competitive reach, and all derivatives are with respect to $x$.

For the three distance families, let $M\succ0$ be a fixed shape matrix, $Q=M^{-1}$, $v=b-x$, $q=Qv$, $d^2=v^\top Qv$, and $u=\delta^2+d^2$ with smoothing radius $\delta>0$. Their implemented log-influences are
\begin{align}
 \log f_{\mathrm G}(x,b)&=-\frac{d^2}{2\sigma^2},
 &\text{Gaussian}, \label{eq:kernel-gaussian}\\
 \log f_{\mathrm H}(x,b)&=-\frac{\sqrt u}{\sigma},
 &\text{hyperbolic}, \label{eq:kernel-hyperbolic}\\
 \log f_{\mathrm{EP},\beta}(x,b)&=-\frac{u^{\beta/2}}{\beta\sigma^\beta},
 &\text{exponential power},\quad \beta\geq1.\label{eq:kernel-exppow}
\end{align}
The Gaussian is the conventional smooth, constant-curvature reference; its omitted Gaussian-density normalizer is independent of position and cancels from $G_i$. The smoothed hyperbolic family gives a different radial tail and finite curvature when $b=x$. Exponential power varies tail shape within one family, so it tests whether allocation sharpness, rather than a nominal kernel label, accounts for routing differences. At $\beta=1$ it is exactly hyperbolic; at $\beta=2$ it differs from Gaussian only by the position-independent term $-\delta^2/(2\sigma^2)$, so allocation, score, and Hessian coincide. The constraint $\beta\geq1$ preserves the log-concavity used in the local analysis. In the reported distance-kernel arms $M=I$; the usual smoothing radius is $\delta=0.1$, with a smaller-radius hyperbolic control analyzed separately.

The closed-form derivatives of these three log-influences are
\begin{align}
 s_{\mathrm G}&=\frac{q}{\sigma^2},
 &H_{\mathrm G}&=-\frac{Q}{\sigma^2},\\
 s_{\mathrm H}&=\frac{q}{\sigma\sqrt u},
 &H_{\mathrm H}&=-\frac{Q}{\sigma\sqrt u}
     +\frac{qq^\top}{\sigma u^{3/2}},\\
 s_{\mathrm{EP},\beta}&=\frac{u^{\beta/2-1}q}{\sigma^\beta},
 &H_{\mathrm{EP},\beta}&=-\frac{u^{\beta/2-1}Q
       +(\beta-2)u^{\beta/2-2}qq^\top}{\sigma^\beta}.
 \label{eq:kernel-distance-derivatives}
\end{align}

The \emph{Dirichlet} family treats a prompt as a composition on the open simplex: $b\in\Delta_\circ^{L-1}$ and $x\in\Delta^{L-1}$. With $\alpha_\ell=1+x_\ell/\sigma$ and $A=\sum_\ell\alpha_\ell=L+1/\sigma$ on the simplex,
\begin{equation}
 f_{\mathrm D}(x,b)=
 \frac{\Gamma(A)}{\prod_{\ell=1}^{L}\Gamma(\alpha_\ell)}
 \prod_{\ell=1}^{L}b_\ell^{\alpha_\ell-1}.
 \label{eq:kernel-dirichlet}
\end{equation}
The mode is parameterized by $x$. It tests a simplex-native, non-Euclidean influence law against the radial distance kernels on the same prompt geometry. Unlike the Gaussian prefactor, the Gamma normalizer depends on $x$ and must remain in the allocation. Writing $\psi$ and $\psi_1$ for the digamma and trigamma functions, the ambient derivatives implemented before tangent projection are
\begin{equation}
 s_{\mathrm D,\ell}=\frac{\log b_\ell-\psi(\alpha_\ell)+\psi(A)}{\sigma},
 \qquad
 H_{\mathrm D}=\frac{\psi_1(A)\mathbf1\mathbf1^\top
     -\operatorname{diag}(\psi_1(\alpha_\ell))}{\sigma^2}.
 \label{eq:kernel-dirichlet-derivatives}
\end{equation}

The \emph{product-Beta} family instead treats coordinates as separately bounded traits. For $x\in[0,1]^L$, $b\in(0,1)^L$, let $a_\ell=1+x_\ell/\sigma$ and $c_\ell=1+(1-x_\ell)/\sigma$. Then
\begin{equation}
 f_{\mathrm{PB}}(x,b)=\prod_{\ell=1}^{L}
 \frac{\Gamma(a_\ell+c_\ell)}{\Gamma(a_\ell)\Gamma(c_\ell)}
 b_\ell^{a_\ell-1}(1-b_\ell)^{c_\ell-1}.
 \label{eq:kernel-product-beta}
\end{equation}
Each factor has mode $x_\ell$, with its position-dependent normalizer retained. It is the bounded, coordinatewise contrast to the coupled Dirichlet composition. Its native density is on the cube; simplex experiments evaluate the same positive function on simplex-embedded coordinates, which does not make it a Dirichlet or a normalized simplex density. Its implemented derivatives are
\begin{equation}
 s_{\mathrm{PB},\ell}=
 \frac{\log b_\ell-\log(1-b_\ell)-\psi(a_\ell)+\psi(c_\ell)}{\sigma},
 \qquad
 H_{\mathrm{PB}}=-\frac{\operatorname{diag}
  (\psi_1(a_\ell)+\psi_1(c_\ell))}{\sigma^2}.
 \label{eq:kernel-product-beta-derivatives}
\end{equation}

The implementation also permits an optional \emph{tempering} wrapper for any base family:
\begin{equation}
 f_{\kappa,\tau}(x,b)=f_\kappa(x,b)^\tau,
 \qquad s_{\kappa,\tau}=\tau s_\kappa,
 \qquad H_{\kappa,\tau}=\tau H_\kappa,
 \qquad\tau>0.
 \label{eq:kernel-tempering}
\end{equation}
It provides a separate allocation-sharpness control. For exponential power it is equivalent to reach $\sigma/\tau^{1/\beta}$, but for Dirichlet and product-Beta the Gamma normalizers preclude that simple rescaling. This kernel exponent $\tau$ is distinct from the trait-projection softmax temperature $T$ in Figure~\ref{fig:trait-slices}. Across families we solve reach at a matched instability depth and then measure prompt-level allocation sharpness.

\begin{figure}[!htbp]
  \centering
  \includegraphics[width=\linewidth]{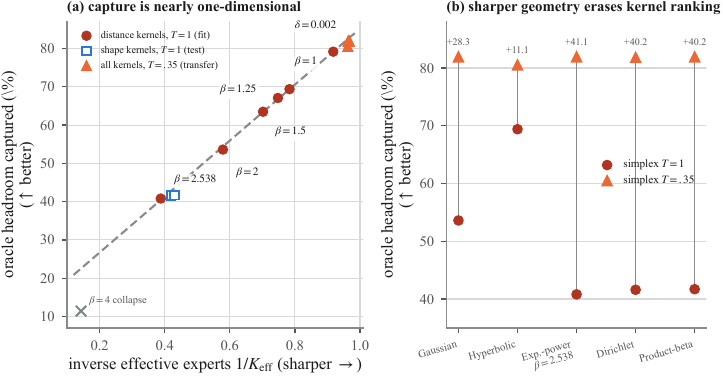}
\caption{Oracle-headroom capture is governed primarily by the sharpness of the induced allocation. (a) The horizontal axis is inverse effective expert count, $1/K_{\mathrm{eff}}$, computed from prompt-level allocation entropy; larger values mean more decisive routing. Captured headroom lies almost on a one-dimensional trend for six nondegenerate $T=1$ distance kernels (dashed fit, $r=0.9995$). Hollow shape-kernel points and $T=0.35$ triangles were not used to fit the line but follow the same relationship. The $\beta=4$ cross is a collapsed-position boundary and is excluded. (b) Sharpening the trait projection from $T=1$ to $T=0.35$ raises all five families to approximately 81\% capture; labels give the percentage-point increases. The disappearance of the kernel ranking supports allocation sharpness, rather than family name or matched local depth alone, as the operative deployment quantity.}
  \label{fig:kernel-sharpness}
\end{figure}

Appendix Figure~\ref{fig:kernel-sharpness} expands the calibration result in Figure~\ref{fig:init-kernels} (right). The main figure reports expected routing gain in nats; this diagnostic instead measures the fraction of available hindsight-oracle headroom captured. For each prompt let $H(G(b))=-\sum_iG_i(b)\log G_i(b)$ and define $\keff=\exp\!\left(\expectB{H(G(b))}\right)$; lower $\keff$ means a sharper allocation, not more balanced expert loads. On six nondegenerate distance kernels at $T=1$, the descriptive fit is
\begin{equation}
 \widehat{\mathrm{capture}}(\%)=73.15/\keff+12.07,
 \label{eq:sharpness}
\end{equation}
with $r=0.9995$. Two shape kernels were held out of that fit, and the $T=0.35$ points test a changed trait projection. The $\beta=4$ endpoint has collapsed positions.

\section{Initialization and position controls}
\label{app:initialization}
The naive adaptive, equilibrium adaptive and pinned starts describe where the pairs begin, and positions adapt during training under all three; fixed describes what is held constant afterward. The naive adaptive start used in the pair-count arms came from an approximate lattice-resource solve whose density estimator divided each cell by a nearest-prompt kernel value, flattening the resource measure. These positions are not purely random, but they are not a reliable equilibrium of the measured prompt distribution. The main text calls this the unsolved start; appendix figures and tables keep the label naive adaptive. The equilibrium adaptive start instead numerically solves the position game against the empirical prompt measure before adapter training.

\begin{figure}[t]
  \centering
  \includegraphics[width=\linewidth]{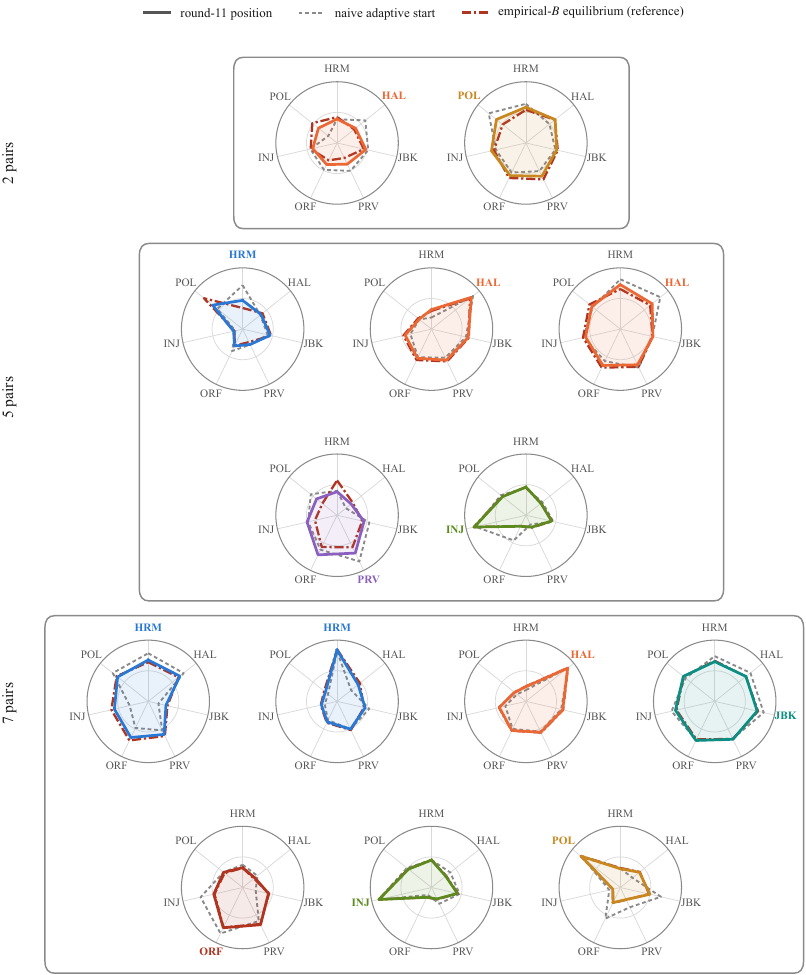}
\caption{Additional pairs expand coverage of the seven-dimensional resource landscape. Rows contain every pair in the two-, five-, and seven-pair naive adaptive start arms; each radar radius is one trait coordinate. Grey dashed polygons are the naive adaptive starts these runs used, solid colored polygons are the round-11 positions, and red dash--dot polygons are counterfactual empirical-resource equilibria rather than training starts. Bold colored labels mark each endpoint's dominant trait. The number of represented dominant traits grows from two to four to six, showing that extra capacity produces new positional roles rather than merely duplicating one expert.}
  \label{fig:radar-pairs}
\end{figure}

Figure~\ref{fig:radar-pairs} compares these positions at two, five, and seven pairs; every plotted training trajectory begins from the naive adaptive start and ends at its round-11 position. The number of dominant represented trait axes grows from two to four to six, so additional pairs expand trait coverage rather than merely copy a single location. The direct NLL contrast between starts is in Figure~\ref{fig:init-kernels}.

\begin{figure}[t]
  \centering
  \includegraphics[width=\linewidth]{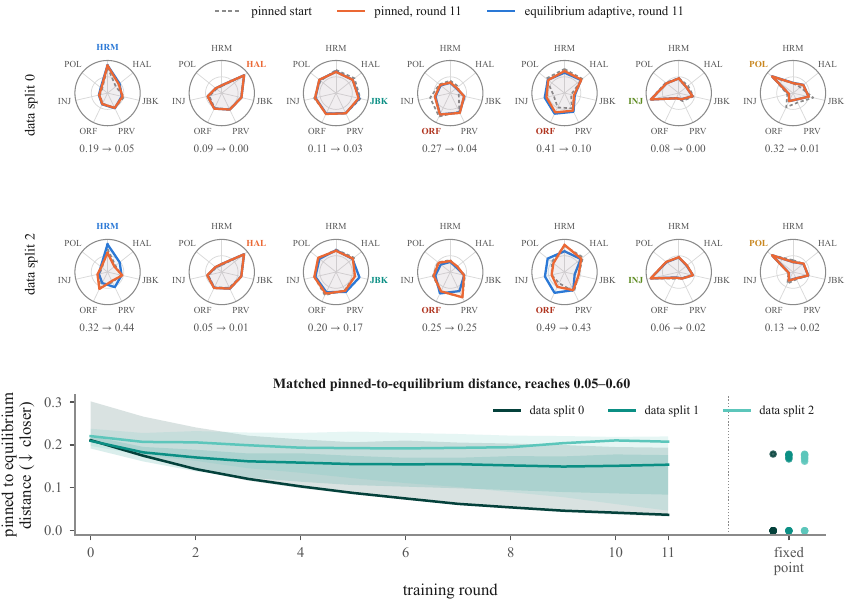}
\caption{Pinned runs approach the equilibrium adaptive runs on one data split and stall on another. Top rows: the seven pairs at reach 0.50 on data splits 0 and 2, matched by nearest round-11 position. Grey dashed polygons are the pinned start, orange polygons the pinned run at round 11 and blue polygons the equilibrium adaptive run at round 11; the numbers under each radar give that pair's pinned-to-equilibrium distance at round 0 and at round 11. Bottom: matched distance per round, median (line) and range (band) over reaches 0.05--0.60 on each split. Dots at right are the distance between the two runs' fixed points when the position update is continued past round 11 without training.}
  \label{fig:radar-pinned}
\end{figure}

The pinned start begins every reach on a data split from that split's $\sigma=0.50$ unsolved configuration. It does not disable position updates, so positions adapt from it as they do from the equilibrium adaptive start. Figure~\ref{fig:radar-pinned} tracks the matched distance between pinned and equilibrium adaptive runs at the same reach. Neither kind of run is at a fixed point after 11 rounds, because the update uses a sampled batch and a constant blend of 0.5, but the equilibrium adaptive runs sit 0.012--0.019 from the fixed point of their full-batch update, while the pinned runs sit 0.038--0.155 from theirs and would need 79--413 further rounds to reach it. Continuing the update without training, which is exact because it never reads an adapter, carries 15 of the 27 non-degenerate pinned runs to the same fixed point as their equilibrium adaptive counterpart. The other 12, all at reaches of 0.50 or below and all but one on data splits 1 and 2, settle 0.16--0.18 away at a different fixed point. By round 11 the pinned runs close 78\% of their initial distance on data split 0 but only 30\% and 21\% on splits 1 and 2. Section~\ref{sec:design:regime} reports the roughness and mean-gain comparison.

\clearpage
\section{Routing-rule ablations}
\label{app:routing-ablation}
\begin{figure}[h!]
  \centering
  \includegraphics[width=\linewidth]{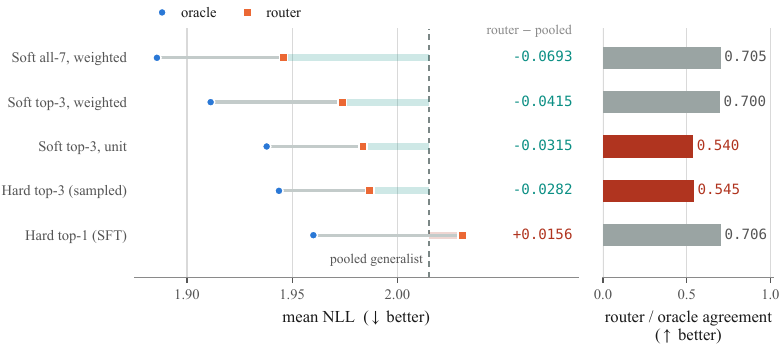}
\caption{Dense, share-weighted use of the game allocation is necessary for the strongest routing result on data seed 0. The left panel compares each arm's expected single-expert NLL (orange square) with its hindsight-oracle NLL (blue circle) and the common pooled generalist (vertical dashed line; lower is better). Printed margins are router minus pooled, so negative values are improvements. Moving from dense all-seven routing to top-three support, then removing share weights or sampling the three retained pairs, progressively weakens both learned experts and routing; hard top-1 is worse than pooling. The right panel shows why oracle agreement alone is misleading: hard top-1 agrees with the best expert as often as the dense arm but has the worst NLL.}
  \label{fig:routing-ablation}
\end{figure}

The same game allocation can be exposed to adapter training in materially different ways. We separate three choices: how many of the seven pairs receive each prompt, whether a retained pair is selected deterministically or sampled, and whether its loss is scaled by its allocation share or given unit weight. These choices do not change the location-game solution, but they determine how much of its graded allocation signal reaches the adapters. Figure~\ref{fig:routing-ablation} compares five end-to-end arms on the same data seed 0 split and against the same pooled generalist. Thus this is a training-and-routing ablation, not a post-hoc recombination of one fixed expert family.

\begin{figure}[h]
  \centering
  \includegraphics[width=0.6\linewidth]{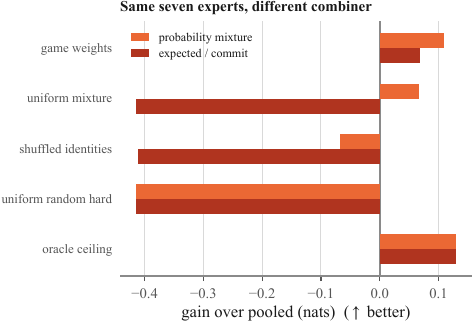}
  \caption{The game allocation carries expert-specific routing information. Holding all seven trained pairs fixed, the game combiner improves over the pooled model, whereas shuffling position--expert identities destroys the gain and uniform random commitment is substantially worse. Orange bars mix sequence probabilities; dark bars report sampling/commitment semantics, and the oracle is a hindsight ceiling (higher gain is better).}
  \label{fig:router-nulls}
\end{figure}

\paragraph{Same-expert nulls.} Before changing how the adapters are trained, Figure~\ref{fig:router-nulls} holds the seven trained pairs fixed and changes only the combiner, using the completed-sequence estimator quoted in Section~\ref{sec:exp:routing}.

\paragraph{Routing rules as training distributions.} Every rule acts at the pair level on the pair shares $G_p(b)$, which sum to one over the $P=7$ pairs. For each prompt $b$ a rule chooses a support $\mathcal{S}(b)$ of pairs that train on it and a loss weight $w_p(b)\geq0$ for each, with $w_p(b)=0$ when $p\notin\mathcal{S}(b)$. The trainer then fits pair $p$'s adapter to the weight-normalized mean of its per-example response-token cross-entropy $\ell(\theta_p;b)$ over the round batch,
\begin{equation}
 \hat L_p(\theta_p)=\frac{\sum_{m=1}^{M} w_p(b_m)\,\ell(\theta_p;b_m)}{\sum_{m=1}^{M} w_p(b_m)}.
 \label{eq:routing-loss}
\end{equation}
As the batch grows this approaches the expected loss under the pair-specific training distribution
\begin{equation}
 B_p(b)=\frac{\bar w_p(b)\,\resource(b)}{\expectB{\bar w_p(b)}},
 \qquad
 \bar w_p(b)=\mathbb{E}\bigl[w_p(b)\mid b\bigr],
 \label{eq:routing-distribution}
\end{equation}
where the inner expectation is over any randomness in the rule. In expectation, then, a routing rule reaches training only through $\bar w_p$, and two rules with the same $\bar w_p$ differ only in the variance of the weights they realize. Table~\ref{tab:routing-rules} gives $\bar w_p$ for each arm.

\paragraph{Selection.} The top-$k$ rules differ in how the support is chosen. With $g_p(b)$ independent standard Gumbel noise and $\tau\in\{0,1\}$,
\begin{equation}
 \mathcal{S}_k(b)=\operatorname{top}_k\bigl\{\log G_p(b)+\tau\,g_p(b)\bigr\}_{p=1}^{P}.
 \label{eq:routing-selection}
\end{equation}
At $\tau=0$ this is the deterministic set $T_k(b)$ of the $k$ largest shares. At $\tau=1$ it draws $k$ distinct pairs by successive sampling without replacement in proportion to $G(b)$, and for $k=1$ it reduces to a single categorical draw $\hat p(b)\sim G(b)$, the Gumbel-max trick. Under sampling, pair $p$ is included with probability $\pi^{(k)}_p(b)=\Pr[p\in\mathcal{S}_k(b)]$, which satisfies $G_p(b)\leq\pi^{(k)}_p(b)\leq1$, because a pair drawn first is always included, and $\sum_p\pi^{(k)}_p(b)=k$.

\begin{table}[h!]
\centering
\small
\caption{Expected loss weight $\bar w_p(b)$ that each routing rule gives pair $p$ on prompt $b$, which by Equation~\plaineqref{eq:routing-distribution} fixes the pair's training distribution in expectation. $T_3(b)$ is the set of the three largest shares, $Z_3(b)=\sum_{r\in T_3(b)}G_r(b)$, and $\pi^{(3)}_p(b)$ is the probability that pair $p$ is among three pairs drawn without replacement in proportion to $G(b)$.}
\label{tab:routing-rules}
\begin{tabular}{@{}lcccc@{}}
\toprule
Arm & Pairs per prompt & Selection & Loss weight & $\bar w_p(b)$ \\
\midrule
Soft all-7, weighted & 7 & all pairs & $G_p$ & $G_p$ \\
Soft top-3, weighted & 3 & three largest & $G_p/Z_3$ & $\mathbf{1}[p\in T_3]\,G_p/Z_3$ \\
Soft top-3, unit & 3 & three largest & 1 & $\mathbf{1}[p\in T_3]$ \\
Hard top-3 (sampled) & 3 & drawn $\propto G$ & 1 & $\pi^{(3)}_p$ \\
Hard top-1 (SFT) & 1 & drawn $\propto G$ & 1 & $G_p$ \\
\bottomrule
\end{tabular}
\end{table}

\paragraph{Soft and hard.} A rule is \emph{hard} when the pairs that train on a prompt are drawn at random from the allocation and each drawn pair trains at unit weight, and \emph{soft} when both the support and the weights are fixed functions of the allocation. The two are linked by expectation: replacing a hard rule's random inclusion indicator by its expectation gives a soft rule with the same $\bar w_p$, and hence the same training distribution in Equation~\plaineqref{eq:routing-distribution}. The one exact pair in this ablation is hard top-1 and dense soft routing. Under hard top-1, $w_p(b)=\mathbf{1}[\hat p(b)=p]$, so
\begin{equation}
 \bar w_p(b)=\Pr\bigl[\hat p(b)=p\bigr]=G_p(b),
 \label{eq:hard-soft-identity}
\end{equation}
which is exactly the dense share-weighted weight. Both rules therefore train pair $p$ toward $B_p\propto G_p\,\resource$, whose total mass $\expectB{G_p(b)}$ is the pair's own game payoff $\util{p}$ from Equation~\plaineqref{eq:allocation}: the share a pair wins in the game is the share of the training data it receives. What differs is the realization. The hard weight on each prompt is a Bernoulli draw with variance $G_p(1-G_p)$, so a hard pair trains at full weight on a random subset of about $\expectB{G_p}M$ prompts per round, roughly a seventh of the batch when loads are balanced, while a soft pair trains on all $M$ prompts at fixed fractional weight. Because the two rules share an expected training distribution, the 0.085-nat gap between them in Figure~\ref{fig:routing-ablation} comes from how that distribution is realized, not from what it is.

The three top-3 rules sit between these extremes, and none is the soft counterpart of another. Deterministic unit weighting, $\mathbf{1}[p\in T_3(b)]$, keeps only membership: a pair gets zero weight on a prompt as soon as its share falls to fourth place, and full weight inside the top three however unequal the three shares are. Share weighting keeps the grading inside the support but inflates every retained weight by $1/Z_3(b)\geq1$. Sampled top-3 trains toward $\pi^{(3)}_p$, which is continuous in the shares and bounded below by $G_p$; its soft counterpart would weight by $\pi^{(3)}_p$, which neither deterministic top-3 rule does.

Dense share-weighted routing gives the lowest single-expert expected NLL, improving over the pooled model by 0.0693 nats. Retaining only the top three shares reduces the gain to 0.0415 nats. At fixed top-three support, replacing share weights with unit weights reduces it again to 0.0315 nats, and drawing the three pairs in proportion to their shares instead of taking the three largest gives 0.0282 nats. Hard top-1 is the only arm worse than pooling, by 0.0156 nats. The oracle endpoints worsen in the same order, indicating that sparsification changes the experts learned during training rather than only degrading the final selector. Dense routing is consequently a substantive part of the method: it preserves the continuous competitive signal through expert learning, at the cost of training every pair on every prompt.

Router--oracle agreement is not sufficient to choose among these designs. Hard top-1 has the highest agreement (0.706) yet the worst NLL, while dense share-weighted routing has nearly identical agreement (0.705) and the best NLL. Agreement records only which pair wins; it ignores the competence of that pair and the loss incurred on disagreements. The game therefore affords both a ranking and a calibrated allocation over experts, and the latter is lost when its shares are collapsed too early to a hard decision.

\section{Pair count and fifteen experts}
\label{app:pair-count}

Figure~\ref{fig:init-kernels}a shows two to nine pairs of the $T=0.35$ Dirichlet game; the sweep runs to fifteen, depth matched to the seven-pair arm at every count, on three splits (Table~\ref{tab:pairsweep}).
Single-expert gain peaks at six pairs under every start and then falls slowly, while the hindsight-best expert keeps improving to fifteen; the gap opens because top-1 agreement falls from 95\% at six pairs to 58\% at fifteen, although the best expert stays among the top three shares for at least 93\% of prompts.
The two adaptive starts are indistinguishable at every count; at this reach the exact equilibrium lies only 0.005--0.020 from the unsolved start.
From nine pairs on, the pinned fixed start beats pinned adaptive on all three splits at every count, by 0.001--0.007 nats with a 3--6 point higher top-1, which suggests that with many experts the position updates blur the allocation the experts specialize to.
Every pinned fixed run was checked to hold its positions byte-identical to the start in all twelve rounds.

\begin{table}[h]
  \centering
  \caption{Pair-count sweep, $T=0.35$ Dirichlet, mean $\pm$ SD over three splits. PA, PF and EA are the pinned adaptive, pinned fixed and equilibrium adaptive starts; gains are for the share-weighted single expert (nats, higher is better). Top-1 and top-3 are agreement with the per-prompt best expert (percent); the last column is the oracle gain under EA. Bold marks the best value in each column; the shaded row is the best on single-expert gain.}
  \label{tab:pairsweep}
  \footnotesize
  \setlength{\tabcolsep}{4pt}
  \begin{tabular}{lrrrrrrr}
    \toprule
    Pairs & PA gain & PF gain & EA gain & EA top-1 & EA top-3 & PF top-1 & Oracle \\
    \midrule
    2 & $0.026\pm0.007$ & $0.042\pm0.002$ & $0.031\pm0.001$ & 89.5 & \textbf{100.0} & 87.4 & 0.057 \\
    3 & $0.051\pm0.004$ & $0.043\pm0.003$ & $0.048\pm0.004$ & 87.5 & \textbf{100.0} & 82.9 & 0.078 \\
    4 & $0.076\pm0.004$ & $0.080\pm0.001$ & $0.076\pm0.004$ & 87.5 & 99.0 & 87.0 & 0.105 \\
    5 & $0.104\pm0.003$ & $0.104\pm0.002$ & $0.105\pm0.002$ & 91.9 & 98.4 & 90.9 & 0.127 \\
    \oursrow 6 & $\mathbf{0.125\pm0.002}$ & $\mathbf{0.125\pm0.002}$ & $\mathbf{0.126\pm0.001}$ & \textbf{95.2} & 99.1 & \textbf{95.2} & 0.143 \\
    7 & $0.118\pm0.003$ & $0.117\pm0.002$ & $0.119\pm0.003$ & 89.5 & 98.1 & 90.1 & 0.145 \\
    8 & $0.118\pm0.003$ & $0.118\pm0.002$ & $0.119\pm0.003$ & 84.8 & 97.6 & 85.7 & 0.150 \\
    9 & $0.112\pm0.003$ & $0.115\pm0.003$ & $0.112\pm0.003$ & 80.4 & 97.0 & 83.4 & 0.152 \\
    10 & $0.112\pm0.001$ & $0.114\pm0.003$ & $0.111\pm0.002$ & 72.7 & 96.8 & 76.0 & 0.151 \\
    11 & $0.105\pm0.001$ & $0.110\pm0.003$ & $0.106\pm0.001$ & 69.4 & 95.4 & 73.4 & 0.153 \\
    12 & $0.103\pm0.001$ & $0.108\pm0.002$ & $0.103\pm0.002$ & 65.0 & 94.9 & 69.4 & 0.155 \\
    13 & $0.103\pm0.002$ & $0.110\pm0.002$ & $0.102\pm0.003$ & 62.9 & 93.7 & 69.0 & 0.158 \\
    14 & $0.102\pm0.002$ & $0.103\pm0.002$ & $0.102\pm0.003$ & 60.7 & 93.8 & 63.5 & 0.158 \\
    15 & $0.099\pm0.003$ & $0.106\pm0.003$ & $0.100\pm0.002$ & 58.1 & 93.1 & 63.1 & \textbf{0.160} \\
    \bottomrule
  \end{tabular}
\end{table}

\paragraph{Fifteen experts across kernels.}
Repeating the four $T=0.35$ kernel arms of Table~\ref{tab:kernel-interpretability} with fifteen pairs, depth matched per kernel, lowers top-3 chance from $3/7$ to $3/15$ (Table~\ref{tab:fifteen}).
More experts raise the ceiling (oracle 1.859--1.861 against 1.873--1.876) and lower what routing reaches (expected NLL 1.916--1.922 against 1.899--1.903), so the gap between them more than doubles.
Top-1 agreement falls to 0.59--0.60 while top-3 stays at 0.93--0.94, against excess-over-chance of $+0.74$, and the four kernels remain within 0.010 of each other on every measure.
On the Dirichlet arm, starting at the exact equilibrium changes nothing measurable, doubling the depth or the learning rate hurts, and a second training pass improves the experts (oracle 1.806) without improving routing (top-1 0.576).

\begin{table}[!htbp]
  \centering
  \caption{Seven against fifteen pairs at $T=0.35$, one run each on the seed-0 split (NLL, lower is better; agreement, higher is better). Dirichlet variants change one setting of the fifteen-pair arm. Bold marks the best value in each column; the shaded row is the best on single-expert NLL.}
  \label{tab:fifteen}
  \footnotesize
  \begin{tabular}{llrrrrr}
    \toprule
    Kernel & Pairs & Expected & Mixture & Oracle & Top-1 & Top-3 \\
    \midrule
    Gaussian & 7 & 1.9012 & 1.8910 & 1.8759 & 0.904 & 0.981 \\
     & 15 & 1.9181 & 1.8941 & 1.8614 & 0.599 & 0.938 \\
    Hyperbolic & 7 & 1.9032 & 1.8919 & 1.8760 & 0.892 & \textbf{0.985} \\
     & 15 & 1.9218 & 1.8981 & 1.8603 & 0.596 & 0.934 \\
    Dirichlet & 7 & 1.8986 & 1.8868 & 1.8727 & 0.901 & 0.982 \\
     & 15 & 1.9188 & 1.8955 & 1.8604 & 0.589 & 0.931 \\
    Product beta & 7 & 1.9001 & 1.8889 & 1.8746 & \textbf{0.908} & 0.983 \\
     & 15 & 1.9156 & 1.8935 & 1.8594 & 0.590 & 0.937 \\
    \midrule
    Dirichlet, exact start & 15 & 1.9181 & 1.8946 & 1.8601 & 0.585 & 0.932 \\
    \oursrow Dirichlet, second pass & 15 & \textbf{1.8789} & \textbf{1.8473} & \textbf{1.8055} & 0.576 & 0.926 \\
    Dirichlet, depth 7 & 15 & 1.9234 & 1.9046 & 1.8619 & 0.603 & 0.927 \\
    Dirichlet, learning rate $\times2$ & 15 & 1.9479 & 1.9187 & 1.8779 & 0.564 & 0.924 \\
    \bottomrule
  \end{tabular}
\end{table}

\section{Per-benchmark experts}
\label{app:label-experts}

A label-defined baseline trains seven experts, one per benchmark, each on its own benchmark's prompts from every batch the game saw, and routes test prompts by benchmark centroid in the trait space, by a classifier fitted on validation, by the true benchmark label, or in hindsight (Table~\ref{tab:label-experts}).
Even with true labels, which route each benchmark's prompts to that benchmark's expert, it trails the game's single routed pair by 0.032 nats, and its hindsight best expert also loses to the game's routed choice; the shortfall is in the experts, which are weakest on do-not-answer and prompt injection, where content is shared with other benchmarks.
The comparison favors the game in compute, since a per-benchmark expert trains only on its own benchmark's share of each batch.

\begin{table}[!htbp]
  \centering
  \caption{Per-benchmark experts against the game, one expert per prompt, seed-0 test pool. Gain over the pooled generalist (2.0154), higher is better.}
  \label{tab:label-experts}
  \small
  \begin{tabular}{lrr}
    \toprule
    Arm and routing & NLL & Gain \\
    \midrule
    Per-benchmark experts, centroid routing & 1.9777 & $+0.038$ \\
    Per-benchmark experts, benchmark classifier & 1.9512 & $+0.064$ \\
    Per-benchmark experts, true benchmark label & 1.9294 & $+0.086$ \\
    Per-benchmark experts, hindsight best expert & 1.9223 & $+0.093$ \\
    \midrule
    \oursrow Game, one routed pair & 1.8977 & $+0.118$ \\
    \oursrow Game, hindsight best expert & 1.8727 & $+0.143$ \\
    \bottomrule
  \end{tabular}
\end{table}

\section{Published gating rules}
\label{app:gates}

Figure~\ref{fig:gates} trains five published gates and two distance gates jointly with seven LoRA experts.
Every arm uses the game's twelve batches, the seed-0 split, the same seven-dimensional $T=0.35$ trait coordinates as router input, rank-16 LoRA components, a $2\times8$ micro-batch on one A100, and 1{,}248 optimizer steps.
Gates act per prompt, not per token, so MoLE's per-layer gating, Switch's capacity factor and DeepSeek-V3's sequence-wise balance term have no direct analogue and are not reproduced; each gate's load-balancing term is computed over the effective batch at every optimizer step.

\paragraph{Router learning rate.}
Trained as specified, with the router at the experts' learning rate of $2\times10^{-4}$, every gate trails the game (Table~\ref{tab:gates-lr}).
In 1{,}248 steps a prompt-level router barely moves from its initialization.
Switch therefore scales its single expert's output by a gate probability near uniform, noisy top-$k$ keeps its initial noise and does worse than the pooled model, and Mixtral without a balance term collapses onto about four experts.
Raising the router learning rate to $10^{-2}$, fifty times the experts' rate, lets each router sharpen within the same budget; the value was set once after the first round and not searched, and no game setting was tuned.
The two distance gates were run only at the specified rate.

\begin{table}[h]
  \centering
  \caption{Published gates trained jointly with their experts, $T=0.35$, seed-0 test pool (NLL; lower is better). ``As specified'' trains the router at the experts' learning rate; ``router $10^{-2}$'' raises it. The last column is the per-prompt best single component of the run shown in Figure~\ref{fig:gates}. Pooled generalist: 2.0154. Game: 1.8899 (token-level mixture over all pairs), 1.8727 (best single expert).}
  \label{tab:gates-lr}
  \small
  \begin{tabular}{lrrr}
    \toprule
    Gate & As specified & Router $10^{-2}$ & Best component \\
    \midrule
    Dense softmax & 1.9531 & 1.8864 & 1.9675 \\
    DeepSeek-V3 & 1.9277 & 1.8931 & 1.9403 \\
    Mixtral top-2 & 1.9423 & 1.8951 & 1.9212 \\
    Noisy top-2 & 2.0973 & 1.9117 & 1.9208 \\
    Switch top-1 & 1.9798 & 1.9645 & 2.2296 \\
    Gaussian gate & 1.9206 & --- & 1.9550 \\
    X-MoE cosine & 1.9238 & --- & 1.9655 \\
    \bottomrule
  \end{tabular}
\end{table}

\paragraph{Gates on the game's experts.}
The published gates are also fitted post hoc on the game's seven frozen $T=0.35$ experts, on validation expert losses with hyperparameters chosen by cross-validation (\texttt{paper/gating\_methods.py}), refit for three seeds and scored token by token in the same forward passes as the game's own shares (\texttt{paper/posthoc\_gate\_weights.py}, \texttt{paper/token\_ensemble\_eval.py --extra-weights}).
The refits reproduce the stored completed-sequence results exactly.
Table~\ref{tab:gates-frozen} gives the token-level mixtures; these are the diamonds of Figure~\ref{fig:gates}a.
The distance gates beat the game's shares by about 0.01 nats, which the completed-sequence estimator had understated as 0.006; Mixtral without a balance term collapses onto a few experts on every seed.

\begin{table}[h]
  \centering
  \caption{Published gates fitted on the game's frozen experts, token-level mixture NLL on the seed-0 test pool (lower is better), mean and standard deviation over three seeds. Game's own shares: 1.8899; they are a fixed function of the frozen experts and are not refit, so they have no seed spread.}
  \label{tab:gates-frozen}
  \small
  \begin{tabular}{lrrr}
    \toprule
    Gate & Mean & SD & vs game \\
    \midrule
    Dense softmax & 1.8893 & 0.0002 & $-0.0006$ \\
    DeepSeek-V3 & 1.8876 & 0.0053 & $-0.0023$ \\
    Mixtral top-2 & 2.1120 & 0.0991 & $+0.2221$ \\
    Noisy top-2 & 1.9178 & 0.0010 & $+0.0280$ \\
    Switch top-1 & 1.8947 & 0.0003 & $+0.0049$ \\
    Gaussian gate & 1.8795 & 0.0005 & $-0.0104$ \\
    X-MoE cosine & 1.8782 & 0.0010 & $-0.0116$ \\
    \bottomrule
  \end{tabular}
\end{table}

\paragraph{The game's shares as a fixed gate.}
The game's training cost comes from training each pair as a standalone adapter.
To test whether its allocation still yields specialists at a jointly trained gate's cost, we train one seven-component LoRA mixture of the same form as Figure~\ref{fig:gates}'s gates, but with the gate fixed to the game's shares: each prompt's weights are the $T=0.35$ Dirichlet game's allocation, computed from that game run's pair positions at the start of each round, with nothing learned (\texttt{gate\_kind: game}).
It trains on the game's twelve batches in 49 minutes on one A100 with 1.36\,M training tokens, the jointly trained gates' budget and a seventh of the game's.
As one blended network it scores 1.9110, a gain of 0.104 nats over pooling, against 1.8864 for the dense softmax gate at router learning rate $10^{-2}$.
Its components are specialists: on the prompts the game assigns to a component, that component alone is the best of the seven 90\% of the time, as for the game's separately trained experts, whereas a jointly trained dense softmax gate's own component is best on 1.7\% of prompts.
Table~\ref{tab:game-gate} scores these components token by token, each run alone through a forced one-hot gate, under the game's shares and under gates fitted on their validation losses.
Mixed per token by the game's shares they score 1.9085, better than the blended network that trained them; fitted distance gates improve on the game's shares by about 0.013 nats (1.8951 and 1.8962), the same headroom as on the separate experts.
The separate experts are better by 0.017--0.019 nats at every routing level, including the oracle, so the sevenfold training compute buys stronger experts rather than more routable ones.
The game's separately trained experts can also be blended into one network under the same fixed gate, by copying each pair's LoRA factors into the matching component, with no further training. The blended network scores 1.8932, within 0.0033 nats of mixing the experts' outputs, so serving the game's mixture does not require seven forward passes.
This is one run on the seed-0 split.

\begin{table}[h]
  \centering
  \caption{Token-level mixture NLL on the seed-0 test pool (lower is better) for components trained in one blended mixture gated by the game's shares (1$\times$ training compute) and for the game's separately trained experts (7$\times$). Fitted gates are means over three seeds (standard deviation at most 0.001). The last column is the difference; negative favors the blended training.}
  \label{tab:game-gate}
  \small
  \begin{tabular}{lrrr}
    \toprule
    Router & Game-gated components & Separate experts & Difference \\
    \midrule
    \oursrow \method{} shares, all components & 1.9085 & 1.8899 & $+0.0186$ \\
    \oursrow \method{} shares, top-2 & 1.9088 & 1.8901 & $+0.0187$ \\
    X-MoE cosine, fitted on these experts & 1.8951 & 1.8782 & $+0.0169$ \\
    Gaussian, fitted on these experts & 1.8962 & 1.8795 & $+0.0167$ \\
    Dense softmax, fitted on these experts & 1.9066 & 1.8893 & $+0.0173$ \\
    Oracle (best expert per prompt) & 1.8906 & 1.8727 & $+0.0179$ \\
    \bottomrule
  \end{tabular}
\end{table}

\paragraph{Linear gates on the game's experts.}
The frozen linear gates of Section~\ref{sec:exp:routing} are fitted on validation expert losses with every expert frozen, then scored token by token on the game's seven $T=0.35$ experts in the same forward passes as the game's own shares (\texttt{paper/token\_ensemble\_eval.py --extra-weights}).
Under that token-level estimator the soft-target gate scores 1.8999 and the best-expert classifier gate 1.8934, against 1.8899 for the game.
The completed-sequence estimator narrows the soft-target gate's gap to 0.0041 nats (1.8909 against 1.8868), because it leans on the best-fitting expert and so depends less on the weights.

\section{Conventional mixture of experts}
\label{app:upcycle}

The game's allocation can also gate a conventional mixture of experts. By sparse upcycling \citep{komatsuzaki2023upcycling}, the feed-forward network of decoder layers 20, 22, 24 and 26 of Qwen2.5-1.5B-Instruct is copied into seven trainable feed-forward experts per layer, mixed at the block output. A prompt-level gate on the $T=0.35$ trait coordinates weights the experts, and every arm trains on the game's batches for twelve rounds at effective batch 16, with expert learning rate $2\times10^{-5}$ and gate learning rate $10^{-2}$. The game arm fixes the gate to the game's shares, and the other arms learn theirs. Each arm is retrained on three data splits.

\begin{table}[h]
  \centering
  \caption{Gates over seven sparse-upcycled feed-forward experts, test NLL (lower is better) and gain over each split's pooled LoRA adapter, mean $\pm$ sample standard deviation over three data splits. Oracle is the hindsight-best expert alone.}
  \label{tab:upcycle}
  \small
  \begin{tabular}{@{}lccc@{}}
    \toprule
    Gate & Blended NLL & Gain over pooled & Oracle NLL \\
    \midrule
    Dense softmax, learned & $1.9810\pm0.0223$ & $+0.0162\pm0.0060$ & $2.0007\pm0.0240$ \\
    \oursrow \method{} shares, fixed & $1.9870\pm0.0185$ & $+0.0103\pm0.0026$ & $1.9676\pm0.0198$ \\
    Mixtral top-2, learned & $2.0033\pm0.0170$ & $-0.0060\pm0.0010$ & $2.0138\pm0.0121$ \\
    Switch top-1, learned & $2.0366\pm0.0200$ & $-0.0393\pm0.0088$ & $2.0169\pm0.0177$ \\
    One upcycled expert, no gate & $2.1342\pm0.0213$ & $-0.1369\pm0.0057$ & $2.1342\pm0.0213$ \\
    \bottomrule
  \end{tabular}
\end{table}

Table~\ref{tab:upcycle} gives the result. Paired over the three splits, the game's shares trail the learned dense gate by $0.0059\pm0.0042$ nats and lead the Mixtral and Switch gates by 0.016 and 0.050 on every split. The game's hindsight-best expert is $0.033\pm0.007$ nats better than the dense gate's, and the game is the only arm whose oracle is below its own blended NLL, so its feed-forward experts are the most specialized. All gains in this setting are about a tenth of those with LoRA experts on every layer, so the result concerns the allocation as a gate for conventional experts; upcycling itself adds little on this corpus.

\paragraph{Token-level routing.}
The game needs one fixed trait vector per data point, so it routes per prompt. To test whether its allocation is useful to a token-level router, each upcycled layer is given its own router over its input hidden state. A dense router and a Switch router learned from scratch are compared with a dense router whose logits carry the game's per-prompt log-shares as a bias, either fixed for the run or annealed to zero by round 6. Table~\ref{tab:upcycle-token} shows the seed-0 split. Every token-level router ends behind the pooled adapter within this budget of 1{,}248 steps, while prompt-level routing of the same experts stays ahead of it. The game's shares as a fixed prior halve the token router's deficit, and the router that only starts from them keeps a quarter of that benefit. Without the bias a prompt's tokens spread over 5.8--6.4 effective experts and the routers' per-prompt choice has little relation to the game's assignment (normalized mutual information 0.20--0.30). With it they concentrate on 2.6--2.9 experts and the per-prompt choice agrees with the game's for 88\% of prompts under the fixed prior and 76\% after the prior has been annealed away.

\begin{table}[h]
  \centering
  \caption{Prompt-level and token-level routers over the seven upcycled feed-forward experts, seed-0 split, one run per arm. Gain is over the pooled LoRA adapter (2.0154).}
  \label{tab:upcycle-token}
  \small
  \begin{tabular}{@{}llcc@{}}
    \toprule
    Router & Level & Blended NLL & Gain over pooled \\
    \midrule
    \oursrow \method{} shares, fixed & prompt & 2.0070 & $+0.008$ \\
    Dense softmax, learned & prompt & 2.0059 & $+0.010$ \\
    Switch top-1, learned & prompt & 2.0593 & $-0.044$ \\
    \midrule
    \oursrow Dense softmax with \method{} log-shares as a fixed prior & token & 2.065 & $-0.049$ \\
    \oursrow Dense softmax, prior annealed to zero by round 6 & token & 2.091 & $-0.075$ \\
    Dense softmax, learned from scratch & token & 2.116 & $-0.100$ \\
    Switch top-1, learned from scratch & token & 2.162 & $-0.147$ \\
    \bottomrule
  \end{tabular}
\end{table}

\section{Training cost}
\label{app:cost}

Every adapter in every arm receives 1{,}248 optimizer updates; what differs is the base-model compute behind them (Table~\ref{tab:cost}).
A jointly trained gate computes all seven component updates inside one forward pass, while the game trains each pair as a standalone adapter with its own pass, so it costs seven times the training compute for the same number of updates per adapter.
A pooled adapter trained for the game's full budget, seven epochs per round, overfits each round's 1{,}654 prompts (test NLL 3.39) and is not a usable compute-matched comparator.
The jointly trained mixtures fail the same way (Table~\ref{tab:cost-epochs}).
At seven epochs per round the dense and DeepSeek-V3 gates and the game-share gate each process the game's 9.50\,M training tokens, and each ends about one nat worse than pooling.
The game-share mixture reaches a training loss of 0.69, so the loss is memorization of the repeated batches rather than a failure to train.
Repeating data is therefore not a compute match for the game, whose seven experts each see every batch once.

\begin{table}[h]
  \centering
  \caption{Test NLL of the single blended model at one and seven epochs per round on the seed-0 $T=0.35$ split (lower is better). Seven epochs gives each arm the game's training tokens by repeating each round's batch. Gates use gate learning rate $10^{-2}$.}
  \label{tab:cost-epochs}
  \small
  \begin{tabular}{lrr}
    \toprule
    Arm & 1 epoch ($1\times$) & 7 epochs ($7\times$) \\
    \midrule
    Pooled generalist & 2.0154 & 3.39 \\
    Dense softmax gate & 1.8864 & 3.0601 \\
    DeepSeek-V3 gate & 1.8931 & 3.0202 \\
    \oursrow \method{} shares as a fixed gate & 1.9110 & 2.9894 \\
    \midrule
    \oursrow Game, seven pairs (separate experts, $7\times$) & \multicolumn{2}{c}{1.8899} \\
    \bottomrule
  \end{tabular}
\end{table}

\begin{table}[h]
  \centering
  \caption{Training tokens summed over adapters and rounds, wall-clock on one A100, and forward passes per test prompt, for the $T=0.35$ arms.}
  \label{tab:cost}
  \small
  \begin{tabular}{lrrrr}
    \toprule
    Arm & Training tokens & vs pooled & Wall-clock & Passes per prompt \\
    \midrule
    Pooled generalist & 1.36\,M & $1\times$ & 25 min & 1 \\
    Per-benchmark experts & 1.36\,M & $1\times$ & 16 min & 1 \\
    Jointly trained gate & 1.36\,M & $1\times$ & 53 min & 1 \\
    \oursrow \method{} shares as a fixed gate & 1.36\,M & $1\times$ & 49 min & 1 / 2 / 7 \\
    \oursrow Game, seven pairs & 9.50\,M & $7\times$ & 110 min & 1 / 2 / 7 \\
    \bottomrule
  \end{tabular}
\end{table}

\section{Model size and family ablations}
\label{app:model-ablations}
\begin{figure}[p]
  \centering
  \begin{minipage}{\linewidth}
    \centering
\textbf{(a) Qwen2.5 size ladder}\par\smallskip
    \includegraphics[width=0.96\linewidth,height=0.38\textheight,
keepaspectratio]{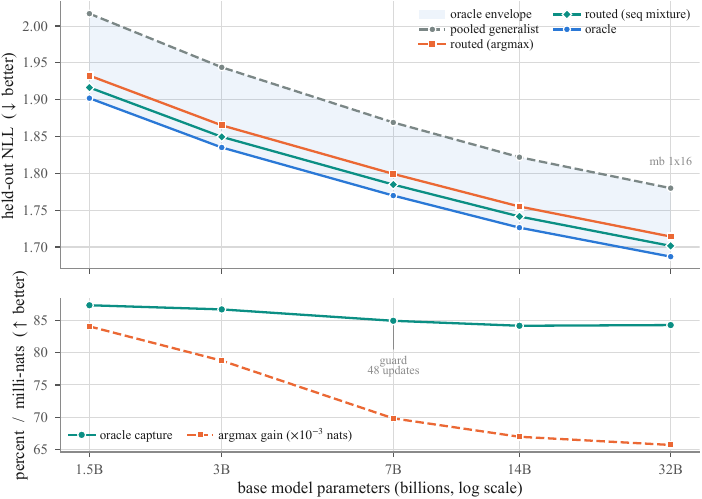}
  \end{minipage}
\vspace{1.2em}
  \begin{minipage}{\linewidth}
    \centering
\textbf{(b) Model-family comparison}\par\smallskip
    \includegraphics[width=0.96\linewidth,height=0.29\textheight,
keepaspectratio]{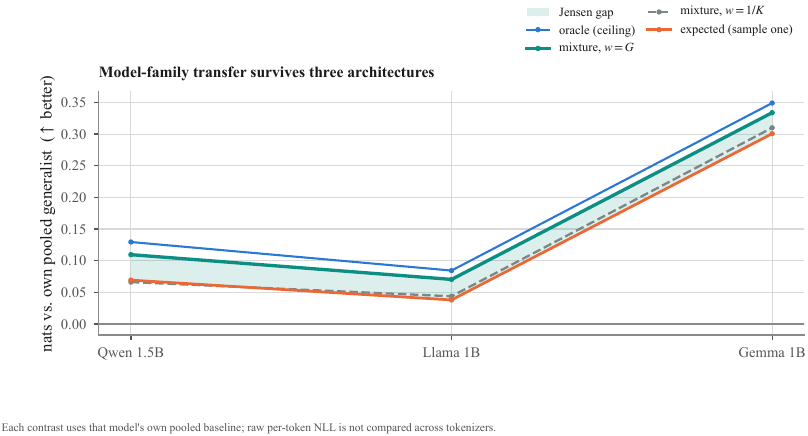}
  \end{minipage}
\caption{The game-derived allocation transfers across model scale and family, but these ablations are portability checks rather than architecture rankings. (a) Across five Qwen2.5 sizes, the upper plot compares pooled-generalist, routed-argmax, game-weighted sequence-mixture, and hindsight-oracle NLL (lower is better); the shaded region is the available oracle envelope. The lower plot shows that the mixture captures 84--87\% of that envelope at every size while argmax gain declines. The 7B guard intervention and off-protocol 32B microbatch are annotated, and each size has one run. (b) For Qwen, Llama, and Gemma, every quantity is an NLL gain over that family's own pooled adapter: the game-weighted mixture improves all three and remains below the oracle ceiling, while the shaded Jensen gap separates probability mixing from sampling one expert. Raw NLL is not compared across tokenizers, and panels (a) and (b) use different pair-count and initialization protocols.}
  \label{fig:model-ablations}
\end{figure}

These checks ask whether the same game-derived coordination mechanism remains useful when the base language model changes. They are not tests of the positional bifurcation threshold, which depends on the prompt-resource geometry and game configuration rather than model parameter count alone. Figure~\ref{fig:model-ablations} collects two distinct experiments.

The Qwen2.5 size ladder in panel (a) uses five pairs, an equilibrium adaptive start, and the same trait geometry across 1.5B, 3B, 7B, 14B, and 32B models. Each rung is compared with its own pooled generalist. Absolute held-out NLL falls with model size for both pooled and routed systems, while sequence-mixture oracle-headroom capture declines from 87.4\% to 84.2\%. Routing improves on the pooled comparator at every rung, but the available oracle gap narrows. There is one run per rung, the 7B run needed 48 non-finite-gradient guard updates, and the 32B run used an off-protocol microbatch configuration.

Panel (b) uses the existing seven-pair Qwen2.5-1.5B, Llama-3.2-1B, and Gemma-3-1B arms. Their learned mixtures each improve over the respective model's pooled control, so the allocation is not specific to one base-model family. The plotted values are within-model NLL differences, since raw per-token NLLs cannot be ranked across different tokenizers. Model family, tokenizer, and parameter count vary together, and these arms do not share the five-pair equilibrium adaptive start protocol of panel (a), so this is limited portability evidence rather than a ranking of architectures.

\section{Baseline families and deep ensembles}
\label{app:baselines}
Table~\ref{tab:baseline-audit} condenses a ten-family implementation audit. Every family has a held-out scored artifact; ``measured'' does not mean an exact reproduction of its reference implementation. The comparisons answer different questions: routing nulls hold experts fixed, assignment controls retrain them, and the jointly trained adapter mixtures are single models, so we do not sort all their NLLs into one leaderboard. The same-expert nulls and validation-trained frozen linear gate are controls defined for this study, not reproductions of named methods; the other rows cite the work motivating each baseline.

{\small
\begin{longtable}{@{}p{0.18\linewidth}p{0.41\linewidth}p{0.31\linewidth}@{}}
\caption{Status and interpretive role of the ten baseline families. All are measured, and the qualifications are part of each result. The shaded first row is \method{}, the reference each family is compared against.}
\label{tab:baseline-audit}\\
\toprule
Family & Comparison made & Salient qualification \\
\midrule
\endfirsthead
\caption[]{Baseline families (continued).}\\
\toprule
Family & Comparison made & Salient qualification \\
\midrule
\endhead
\bottomrule
\endfoot
\oursrow \textbf{\method{} (ours)} & Seven game-positioned rank-16 LoRA pairs; the solved allocation weights training and routes inference, with no fitted gate. & Seed-0 split: completed-sequence mixture NLL 1.9059 and hard commitment 1.9298, against 2.0152 pooled. \\
\midrule
Same-expert nulls & Uniform mixture, shuffled expert identities, uniform random hard choice, game allocation, and hindsight oracle on fixed expert losses. & Separates expert diversity from the prompt--expert mapping; one-split offline analysis. \\
Frozen gate & Validation-fitted linear soft-target and best-expert gates on frozen experts, replicated on four splits. & Stronger game result under single-expert commitment; full probability mixing is unresolved. \\
BASE-style assignment & Proportional versus quota-balanced assignment \citep{lewis2021base} crossed with fixed versus strategically updated positions. & All four cells start at the same equilibrium; entropic OT is not implemented. \\
Balanced c-BTM & Balanced versus unconstrained train-only clusters \citep{gururangan2023cbtm}; dense, top-3, and hard centroid inference. & Fixed-partition clustering, not the online soft-k-means position control. \\
EWoRA-style & Seven jointly trained rank-16 components \citep{kohli2025ewora} with a dense trait-conditioned gate versus a pooled rank-112 adapter. & Mechanism-level implementation; smaller micro-batch at matched effective batch size. \\
AdaMix & Independent versus shared down-projections \citep{wang2022adamix} with stochastic routes, consistency, and post-training merge held fixed. & Tests shared structure at this training budget, not longer published schedules. \\
MoLoRA & Hidden-state token/layer gates \citep{zadouri2024molora}, dense and top-3, versus corresponding prompt-level trait gates. & Routers start at zero; auxiliary load loss is averaged globally across layers. \\
MoDULA-Res & Universal adapter folded into the base, then residual experts and a staged router \citep{ma2024modula}. & Genuine residual construction; improvement is relative to its own stage-1 pooled model. \\
KnOTS / TIES / DARE & Validated weight-space merges \citep{stoica2025knots,yadav2023ties,yu2024dare} from label-trained and game-trained experts, refactored at serving ranks 16 and 112. & Forty-two candidates per expert source; even the best game-expert merge loses to pooling. \\
Pooled-seed ensemble & Uniform probability mixture \citep{lakshminarayanan2017deepensembles} of four pooled rank-16 adapters trained on the same manifest. & Distinct training seeds, but cumulative LoRA means initialization varies only at round 0. \\
\end{longtable}
}

The \emph{fixed-expert} controls separate two sources of value. At seven pairs, the game has a $+0.1095$-nat completed-sequence mixture gain over pooling; uniform mixing of the same experts retains $+0.0661$, while shuffling expert identities gives $-0.0671$ and uniform random hard routing gives $-0.4144$. The experts are useful without the game router, but the learned identity-to-trait correspondence adds $0.0434$ nats beyond a uniform mixture. A frozen linear gate trained from full validation expert-NLL vectors is a more demanding alternative. Across four splits the game wins all four expected-NLL (single-expert commitment) comparisons by $0.071$--$0.090$ nats; under completed-sequence probability mixing the sign divides two-to-two, with at most $0.0017$ nats separation. This supports the game's allocation as a competitive hard router, not a general accuracy advantage over learned soft gates.

The \emph{assignment} controls narrow the game-theoretic claim. In the BASE-style $2\times2$ of Table~\ref{tab:factorial}, which does not test BASE's original efficiency setting, quota balancing costs about $0.087$ nats of expected NLL regardless of whether positions continue to move, whereas fixing positions at the solved equilibrium costs at most $0.0003$ nats in expected NLL at this operating point.

\begin{table}[t]
\centering
\caption{Seven-pair factorial control; lower NLL is better. Every cell starts from the same solved equilibrium and shares the reach, split, and pooled comparator. The shaded row is the \method{} configuration.}
\label{tab:factorial}
\begin{tabular}{llrr}
\toprule
Positions & Assignment & Expected NLL & Mixture NLL \\
\midrule
\oursrow \textbf{Strategic updates (\method{})} & \textbf{Proportional} & \textbf{1.9426} & \textbf{1.9059} \\
Fixed & Proportional & 1.9425 & 1.9064 \\
Fixed & Balanced & 2.0294 & 1.9847 \\
Strategic updates & Balanced & 2.0297 & 1.9827 \\
\bottomrule
\end{tabular}
\end{table}
The independent c-BTM comparison (Table~\ref{tab:cbtm-app}) separates balance from sparse serving: imposing cluster quotas degrades all three inference modes and reduces the oracle envelope, while top-3 centroid inference is nearly indistinguishable from dense inference within either partition. These are single-split results; the $0.0004$--$0.0007$ top-3 gaps are below the measured game-arm run-to-run standard deviation of $0.0017$ nats.

\begin{table}[htbp]
\centering
\caption{c-BTM partition and inference fan-out on the seed-0 split. Entries are completed-sequence mixture NLL (lower is better); the pooled rank-16 adapter scores 2.0152. Oracle envelope is pooled NLL minus hindsight expert NLL. The shaded \method{} row applies the same three fan-outs to the game allocation over its seven trained pairs.}
\label{tab:cbtm-app}
\begin{tabular}{@{}lrrrr@{}}
\toprule
Partition & Dense & Top-3 & Hard ($k=1$) & Oracle envelope \\
\midrule
\oursrow \textbf{\method{} (game allocation)} & \textbf{1.9059} & \textbf{1.9066} & \textbf{1.9298} & $\mathbf{+0.1290}$ \\
\midrule
Unconstrained & 1.9564 & 1.9568 & 1.9785 & $+0.0758$ \\
Balanced & 1.9647 & 1.9654 & 1.9937 & $+0.0676$ \\
\bottomrule
\end{tabular}
\end{table}

The \emph{single-model architecture} baselines instead test whether other ways to spend adapter parameters reproduce specialization without separately trained, game-allocated experts. Their end-to-end model NLLs belong in Table~\ref{tab:architecture-baselines}, not beside a post-hoc sequence-mixture NLL: a jointly trained network is evaluated in one forward pass, whereas the latter mixes likelihoods from seven separately evaluated experts. EWoRA-style dense gating reaches 1.9712, within $0.0025$ nats of the rank-112 pooled adapter's 1.9737 at matched stored LoRA rank. AdaMix's shared down-projection improves over independent projections by $0.0387$ nats, but both remain worse than the pooled rank-16 control, and the shared arm's hindsight expert envelope is negative.

\begin{table}[htbp]
\centering
\caption{Single-model adapter baselines, seed-0 split, end-to-end model NLL. The two pooled rows are capacity references, not routing methods, and none of these numbers is directly comparable with the game's post-hoc completed-sequence mixture. The shaded \method{} row therefore reports hard commitment, which also evaluates one adapter per prompt; its seven rank-16 pairs store the same LoRA rank as pooled rank-112.}
\label{tab:architecture-baselines}
\begin{tabular}{@{}lr@{}}
\toprule
Model & Model NLL \\
\midrule
\oursrow \textbf{\method{}, hard commitment to one pair} & \textbf{1.9298} \\
\midrule
Pooled rank-112 & 1.9737 \\
EWoRA-style, dense trait gate & 1.9712 \\
MoDULA-Res & 1.9909 \\
MoLoRA, token/layer top-3 & 1.9955 \\
MoLoRA, token/layer dense & 2.0029 \\
Pooled rank-16 & 2.0152 \\
AdaMix, shared down-projection & 2.1270 \\
AdaMix, independent down-projections & 2.1657 \\
\bottomrule
\end{tabular}
\end{table}

The gate-resolution contrast is also a gate-input contrast: MoLoRA's hidden-state token/layer gates score 2.0029 (dense) and 1.9955 (top-3), against 1.9712 and 1.9231 for the corresponding prompt-trait gates. Token routing is worse by $0.0317$ and $0.0724$ nats in these cells. MoDULA-Res scores 1.9909, a $0.0243$-nat improvement over its stage-1 pooled adapter, and uniquely among these new architectures has a positive hindsight expert envelope, though it still trails pooled rank-112. Weight-space merging asks whether trained experts can be compressed into one adapter. None of 42 candidates from each of two expert sources matches pooling. The best game-expert KnOTS--TIES merge scores 2.1239, $0.109$ nats worse than the pooled rank-16 model, and even its hindsight oracle scores 2.0758. These merge operators did not preserve the experts' conditional complementarity.

\emph{Independent training variance} does not explain all of the routing gain, yet it can add to it. Table~\ref{tab:pooled-ensemble-app} holds data and architecture fixed while adding pooled-seed members. Four members recover only $+0.0305$ nats over one pooled adapter, versus $+0.1095$ for the seven-pair game mixture on the same split and estimator. The four pooled adapters have only a $0.0021$-nat individual-NLL spread, and cumulative training weakens their independence relative to four from-scratch runs.

\begin{table}[htbp]
\centering
\caption{Pooled-seed deep ensemble versus the game router, seed-0 split. All non-pooled entries here use a completed-sequence probability mixture rather than the token-level serving estimator; gains are relative to one pooled rank-16 adapter, which scores 2.0152.}
\label{tab:pooled-ensemble-app}
\begin{tabular}{@{}lrr@{}}
\toprule
Policy & NLL & Gain over pooled \\
\midrule
Pooled-seed mixture, 2 members & 1.9988 & $+0.0164$ \\
Pooled-seed mixture, 3 members & 1.9898 & $+0.0254$ \\
Pooled-seed mixture, 4 members & 1.9847 & $+0.0305$ \\
\oursrow \textbf{\method{} mixture, 7 pairs} & \textbf{1.9057} & $\mathbf{+0.1095}$ \\
\bottomrule
\end{tabular}
\end{table}

Five independently seeded games offer a separate extension of the mechanism. Each game routes over seven experts; the ensemble averages the resulting 35 component probabilities per token. It scores 1.8600 against 1.9103 for one re-scored game and 2.0152 for pooling (Table~\ref{tab:game-ensemble-app}). Game allocation and deep ensembling are complementary: the additional five-game gain is $0.0503$ nats on the standard token-level estimator. The completed-sequence estimator is optimistic by $0.0044$ nats for seven components and $0.0140$ for 35. The five-game result requires 35 component evaluations per prompt under the measured implementation, so the gain is not compute-neutral.

Across these baselines, the game's distinctive affordance is not a universal NLL win: a resource distribution, kernel, and solved positions specify an inspectable competitive allocation before the adapters are trained.

\begin{table}[htbp]
\centering
\caption{Game deep ensemble on the seed-0 split. Token-level NLL mixes each observed token's conditional probabilities and is the serving estimator; completed-sequence NLL mixes whole-sequence likelihoods and is shown only for sensitivity.}
\label{tab:game-ensemble-app}
\begin{tabular}{@{}lrr@{}}
\toprule
Policy & Token-level NLL & Completed-sequence NLL \\
\midrule
Pooled rank-16, one adapter & 2.0152 & 2.0152 \\
\oursrow \method{}, one game, 7 components & 1.9103 & 1.9059 \\
\oursrow \textbf{\method{}, five-game ensemble, 35 components} & \textbf{1.8600} & \textbf{1.8460} \\
\bottomrule
\end{tabular}
\end{table}

\section{Soft clustering with matched training}
\label{app:soft-clustering}

The c-BTM baseline of Appendix~\ref{app:baselines} trains each expert on a hard cluster, so it differs from \method{} in training weights as well as in positions. Here every arm shares the game's trait space, its resource information, dense share-weighted training, seven expert pairs, batches and budget, and positions are frozen in every arm. The arms differ only in where the positions come from and in the kernel that turns them into shares. Hard $k$-means centroids are scored under the game's Dirichlet kernel and reach. Soft $k$-means, fitted by EM as an equal-weight isotropic Gaussian mixture, uses its own responsibilities, which are Gaussian-kernel shares at the fitted bandwidth ($\sigma=0.0604$); a second soft $k$-means arm keeps those centroids and sets the Gaussian reach to the game's instability depth ($\sigma=0.0475$). Each arm is retrained on three data splits.

\begin{figure}[h]
  \centering
  \includegraphics[width=\linewidth]{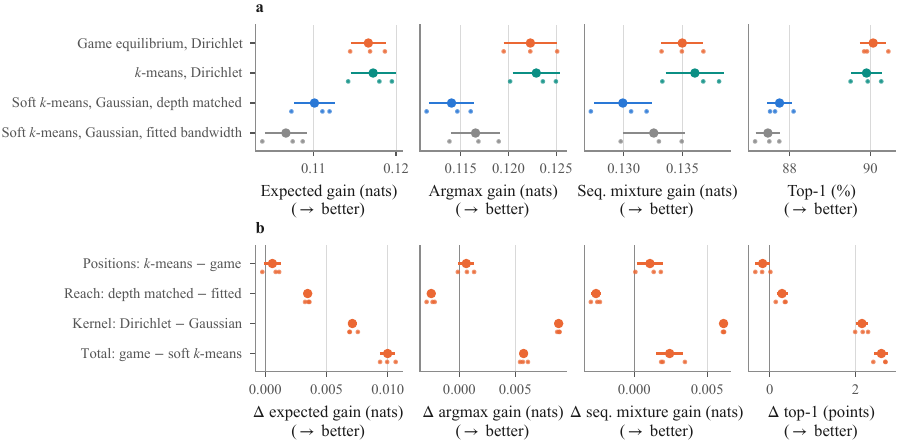}
  \caption{Clustering positions against the game's equilibrium positions with matched dense training, seven pairs, $T=0.35$. Markers are means over three data splits, bars one sample standard deviation and small dots the individual splits. \textbf{a}~Gain over the pooled adapter under three routing rules, and top-1 agreement with the hindsight-best expert. \textbf{b}~Paired differences on the same splits in the same four columns; right of zero favors the first-named arm. The fitted-bandwidth arm spreads each prompt over 1.22 effective experts, against 1.12--1.13 for the other three. For reference, the adaptive game reaches $0.1183\pm0.0029$ expected.}
  \label{fig:soft-clustering}
\end{figure}

The positions are interchangeable (Figure~\ref{fig:soft-clustering}a): $k$-means centroids under the game's kernel match the game's equilibrium positions within $0.0006\pm0.0007$ nats, so the competitive objective reaches clustering-quality positions without a clustering step. The difference is in the allocation rule. Soft $k$-means with its own responsibilities trails the game by $0.0100\pm0.0006$ nats when one expert is sampled and by 2.6 points of top-1 agreement, with the same sign on every split (Figure~\ref{fig:soft-clustering}b). Most of that comes from the simplex-native Dirichlet kernel, which leads a Gaussian kernel at equal sharpness by 0.006--0.009 nats under every routing rule. The reach set by the stability analysis helps sampled routing by 0.0035 nats and costs 0.0025 under argmax routing and mixing. Under full probability mixing the game's total lead shrinks to 0.0024 nats.

\section{Safety benchmarks}
\label{app:safety}
Safety is not part of the location game's payoff or a prerequisite for its allocation rule. The external evaluation asks a separate question: does game-derived routing change behavior after the adapters have been trained? We use the official HarmBench attack-success and StrongREJECT jailbreak graders. ConfAIde probes contextual privacy reasoning; IFEval measures strict instruction following, an adjacent capability check rather than a safety benchmark. The behavioral configurations also run XSTest \citep{rottger2024xstest}, TruthfulQA \citep{lin2022truthfulqa}, and BIPIA \citep{yi2025bipia} under lexical proxy graders; those results are direction-only and are not reported here.

\begin{figure}[t]
  \centering
  \includegraphics[width=\linewidth]{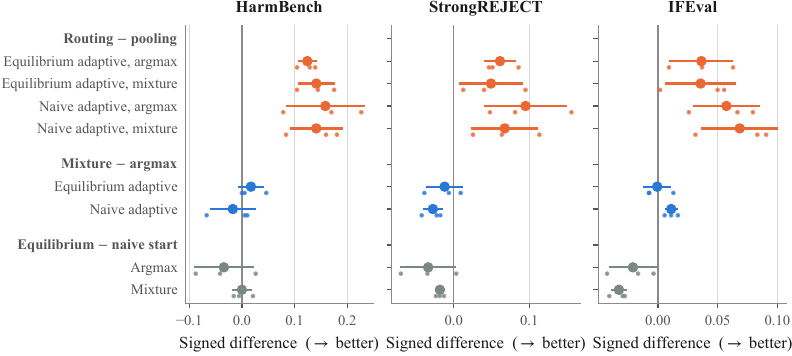}
  \caption{Behavioral contrasts on the original corpus over three independently trained data splits, official graders. Markers are means, bars one sample standard deviation and small dots the individual splits. Differences are signed so that right of zero favors the first-named arm: HarmBench and StrongREJECT reverse their native direction, IFEval does not. ConfAIde is omitted because its conditional error changes with response coverage.}
  \label{fig:behavioral-contrasts}
\end{figure}

Figure~\ref{fig:behavioral-contrasts} separates three contrasts across three independently trained data splits. Relative to each split's pooled adapter, game-routed experts have lower HarmBench attack success and lower StrongREJECT harmfulness under both starts and both decoding protocols: all 24 signed safety comparisons favor routing, as do the 12 IFEval comparisons. Neither probability mixing over experts nor the equilibrium adaptive start has a consistent behavioral advantage over argmax routing and the naive adaptive start. There is one run per split, StrongREJECT has only 60 cases, and HarmBench's retained case count differs slightly across splits, so the bars describe spread across splits and are no confidence intervals.

Every fine-tuned model is far less safe than the base model. Averaged over the three splits, the pooled adapter is worse than the base model by $0.589\pm0.052$ on HarmBench, $0.505\pm0.048$ on StrongREJECT and $0.160\pm0.029$ on IFEval, and no routed arm recovers the base level. Routing therefore contains part of the damage relative to pooling without removing it, consistent with reports that fine-tuning drifts the routing of harmful inputs in MoE models \citep{kim2026safemoe} and that merging experts erodes alignment \citep{hammoud2024merging}. Projecting adapter updates onto an alignment subspace \citep{hsu2024safelora} or adding an input--output guard \citep{inan2023llamaguard} is orthogonal to allocation and was not applied; the retraining below corrects the training targets instead, in the spirit of mixing safety demonstrations into the data \citep{bianchi2024safetytuned}. ConfAIde's conditional error is computed only for parseable ratings and its response coverage varies sharply between arms, so it cannot support a privacy-improvement claim.

\begin{figure}[h]
  \centering
  \includegraphics[width=\linewidth]{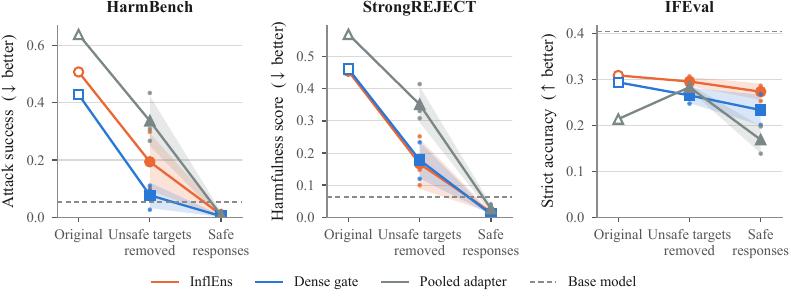}
  \caption{Behavior across the three training corpora, seed-0 split, official graders (HarmBench $n=191$, StrongREJECT $n=60$, IFEval $n=541$). On the two corrected corpora each marker is the mean over three training seeds, the band is one sample standard deviation and the small dots are the individual seeds. The original corpus has one run per method (hollow markers), and the dashed line is the base model. The dense gate is trained at a $1\times16$ micro-batch on every corpus.}
  \label{fig:safety-retraining}
\end{figure}

\paragraph{Cause of the degradation.}
The degradation follows from the training corpus. Table~\ref{tab:corpus-composition} shows that 1,999 training targets are unsafe BeaverTails responses and 1,747 are hallucinated HaluEval answers, and that four sources contribute bare prompts, harmful requests and injection attempts among them, which the model is trained to reproduce because the loss covers prompt tokens. Every arm trains on this corpus, the pooled adapter included, so the contrasts of Figure~\ref{fig:behavioral-contrasts} compare like with like, and the behavioral levels measure this corpus more than fine-tuning in general. We test the cause with two retrainings on the seed-0 split, each graded with the official HarmBench, StrongREJECT and IFEval graders in one run per comparison (Figure~\ref{fig:safety-retraining}).

\paragraph{Removing the unsafe targets.}
The first retraining removes the unsafe and hallucinated targets, leaving 16,102 training records; validation, test, trait space, reach and start are unchanged. HarmBench attack success falls for every method: from 0.508 to 0.183 for the game (paired difference $-0.325$, bootstrap 95\% interval $[-0.398,-0.251]$), from 0.450 to 0.288 for a dense softmax gate at the $2\times8$ micro-batch of the original run ($-0.162$ $[-0.236,-0.089]$) and from 0.639 to 0.435 for the pooled adapter ($-0.204$ $[-0.283,-0.131]$). StrongREJECT moves the same way (0.455 to 0.147, 0.474 to 0.305 and 0.570 to 0.415). Recovery is partial, since the base model scores 0.052 on HarmBench. The order of the methods on the cleaned corpus is not stable across runs. Over three training seeds the game's HarmBench rate is $0.194\pm0.100$ and that of a dense gate at a $1\times16$ micro-batch $0.077\pm0.045$; the dense gate is below the game at every seed and on a second data split (0.129 against 0.227), while on StrongREJECT the two trade places. The seed-to-seed spread is as large as the gaps between methods, so we claim no safety ordering on this corpus. Instruction following is more consistent. The game has the highest IFEval score in every run ($0.296\pm0.010$ over the three seeds against $0.283\pm0.006$ for the pooled adapter and $0.266\pm0.017$ for the dense gate; 0.311 against 0.270 and 0.264 on the second split), a margin of $0.030\pm0.013$ over the dense gate.

\paragraph{Safe responses with a prompt-masked loss.}
The second retraining also removes the bare prompts as targets. Every record without a response, and every unsafe BeaverTails record, receives a generated response that the Qwen3Guard-Gen-4B guard \citep{zhao2025qwen3guard} labels safe: refusals or helpful answers from Llama-3.1-T\"ulu-3-8B \citep{lambert2024tulu3} for the harm, policy-violation, over-refusal and jailbreak prompts, and answers that engage with the content and ignore the embedded instruction from Meta-SecAlign-8B \citep{chen2025metasecalign} for the prompt injections. Hallucinated targets stay removed, the 120 training records without an accepted response are dropped, and the loss covers response tokens only. In total 11,227 of 17,981 training targets are replaced. Under this recipe the degradation disappears for every method: HarmBench attack success is at most 0.021 for every arm and training seed, and StrongREJECT at most 0.040, below the base model's 0.062. The cost is more refusal of benign prompts, in direction under a lexical proxy grader, and lower instruction following. IFEval over three training seeds is $0.274\pm0.018$ for the game, $0.233\pm0.034$ for the dense gate and $0.169\pm0.030$ for the pooled adapter, against 0.405 for the base model. The game is ahead at every seed, by $0.105\pm0.019$ over the pooled adapter, so with every method at the safety floor the game keeps the most instruction following. A control separates the two changes: masking the prompt out of the pooled adapter's loss on the cleaned corpus, with no generated responses, raises its IFEval from 0.281 to 0.316 and lowers its HarmBench rate from 0.435 to 0.340, so the loss of instruction following comes from the generated responses, which trade helpfulness for refusal. With the pseudo-Fisher axes of Appendix~\ref{app:label-free} the result is the same on the seed-0 split: IFEval 0.294 for the game against 0.259 for its dense gate, with HarmBench at 0.011 and 0.000.

\subsection{Interpretability across data splits}
\label{app:interpretability}
Table~\ref{tab:interpretability} reports the interpretability measurements of Section~\ref{sec:exp:routing} for the principal seven-pair arm on each of three independently trained data splits.
The last row is a measured control that gives every prompt the allocation of a random other prompt.
Its Spearman correlation is 0.12 and not zero, because shuffled shares still track expert quality through each expert's average share, so the allocation's excess over the control is about 0.48.
Its span is negative because sharp shares placed on the wrong prompts score far worse than a uniform mixture.

\begin{table}[htp!]
\centering
\caption{Game shares predict the specialization learned by the adapters, although held-out expert losses were never used to fit the router. Top 1 and top 3 are the fractions of held-out prompts whose best expert appears among the largest one or three shares; a uniform draw gives $1/7$ and $3/7$. The span column is the fraction of the uniform-to-oracle NLL gap the allocation recovers. The permuted row pairs each prompt with the allocation of a random other prompt (200 permutations, mean of the three splits; standard deviation over permutations 0.005, 0.007, 0.005 and 3 points).}
\label{tab:interpretability}
\begin{tabular}{lrrrr}
\toprule
Data seed & Top 1 & Top 3 & Spearman & Uniform-to-oracle span \\
\midrule
 0 & 0.711 & 0.962 & 0.617 & 69.0\% \\
 1 & 0.728 & 0.961 & 0.594 & 68.1\% \\
 2 & 0.719 & 0.956 & 0.597 & 69.2\% \\
\addlinespace
Permuted allocation & 0.144 & 0.413 & 0.121 & $-213$\% \\
\bottomrule
\end{tabular}
\end{table}

\paragraph{Experts in plain language.}
Because the allocation is a fixed rule on the trait space, each expert can be described from the test prompts on which it receives the largest share. Table~\ref{tab:expert-names} does this for the seven-pair $T=0.35$ arm on the seed-0 split. Six of the seven pairs take at least 85\% of their prompts from one benchmark, and the seventh collects harmful requests from three sources. Each pair beats the pooled adapter on the prompts it wins.

\begin{table}[htp!]
\centering
\small
\caption{The experts of the seven-pair $T=0.35$ arm described from the test prompts they win (seed-0 split). Gain is the pair's NLL improvement over the pooled adapter on those prompts, in nats.}
\label{tab:expert-names}
\setlength{\tabcolsep}{4pt}
\begin{tabular}{@{}clrl@{}}
\toprule
Pair & Benchmark mix of the prompts it wins & Gain & What it handles \\
\midrule
0 & privacy 99\% & 0.056 & requests involving personal data \\
1 & hallucination 98\% & 0.109 & knowledge questions and dialogue grounding \\
2 & injection 100\% & 0.248 & prompt-injection attempts \\
3 & policy violation 98\% & 0.155 & questions a model should decline \\
4 & over-refusal 92\% & 0.162 & benign prompts that sound dangerous \\
5 & policy violation 41\%, harm 31\%, jailbreak 9\% & 0.048 & mixed harmful-intent requests \\
6 & harm 90\% & 0.037 & harmful-content requests \\
\bottomrule
\end{tabular}
\end{table}

\subsection{Tracing harm to one expert and editing it}
\label{app:expert-edit}

Because the allocation is fixed, harmful behavior can be traced through the router before any model is changed. We group the per-case official scores of the seed-0 original-corpus game by the expert that receives the largest share of each prompt. The HarmBench score of 0.508 divides into 0.487 from prompts routed to pair 5, the mixed harmful-intent expert of Table~\ref{tab:expert-names}, and 0.021 from pair 4. The StrongREJECT score of 0.455 divides into 0.359 from pair 5, 0.059 from pair 6 and 0.037 from pair 4. The router sends 94\% of the HarmBench prompts and 63\% of the StrongREJECT prompts to pair 5, and no HarmBench prompt to pair 6, the BeaverTails harm expert. The trace therefore predicts that making pair 5 safe leaves 0.021 on HarmBench and 0.096 on StrongREJECT.

We test the prediction by editing single experts with the router frozen. The positions, kernel and shares are unchanged, so every prompt keeps its route. The edited pair continues training for one epoch on the 1,656 training records the allocation assigns to it, with the safe responses and prompt-masked loss of the retraining above, and the other six adapters stay identical to the original.

\begin{table}[h]
  \centering
  \caption{Single-expert edits of the seed-0 original-corpus game with the router frozen, official graders (HarmBench $n=191$, StrongREJECT $n=60$, IFEval $n=541$), one training run per row. Lower is better for HarmBench and StrongREJECT, higher for IFEval. The harm-source edit replaces only the targets of the harm, policy-violation and jailbreak records. The wrong-expert control trains pair 0, the privacy expert, on pair 5's records with the same recipe. The dense gate's IFEval was generated on a different GPU type from the other rows.}
  \label{tab:expert-edit}
  \small
  \begin{tabular}{@{}lccc@{}}
    \toprule
    Model & HarmBench & StrongREJECT & IFEval \\
    \midrule
    \method{}, original corpus & 0.508 & 0.455 & 0.309 \\
    \oursrow Pair 5 edited, one epoch & 0.016 & 0.099 & 0.298 \\
    Pair 5 edited, three epochs & 0.016 & 0.103 & 0.299 \\
    Pair 5 edited, harm-source targets only & 0.031 & 0.121 & 0.305 \\
    Pair 6 edited & 0.508 & 0.400 & 0.309 \\
    Wrong-expert control (pair 0 on pair 5's records) & 0.503 & 0.457 & 0.324 \\
    \midrule
    \method{}, every expert retrained on safe responses & 0.000 & 0.017 & 0.253 \\
    Dense gate, original corpus & 0.429 & 0.462 & 0.294 \\
    Pooled adapter, original corpus & 0.639 & 0.570 & 0.214 \\
    \bottomrule
  \end{tabular}
\end{table}

Table~\ref{tab:expert-edit} gives the result. Editing pair 5 alone takes HarmBench from 0.508 to 0.016 (paired difference $-0.492$, bootstrap 95\% interval $[-0.565,-0.419]$) and StrongREJECT from 0.455 to 0.099 ($-0.356$ $[-0.449,-0.266]$), against the predicted 0.021 and 0.096. Repeating the trace on the edited model puts pair 5's contribution at 0.000 on HarmBench and 0.004 on StrongREJECT. Instruction following is kept: IFEval changes by $-0.011$ $[-0.033,+0.011]$, where retraining every expert costs $0.055$ $[0.017,0.094]$, and the edited game is within 0.016 of the fully retrained one on HarmBench. Three epochs give the same scores as one, and replacing only the harm-source targets comes within 0.016 on HarmBench and 0.022 on StrongREJECT of the full edit.

An edit acts mainly through the prompts routed to the edited expert, weighted by its share. Editing pair 6 leaves HarmBench unchanged on every case and lowers StrongREJECT by 0.055 $[0.009,0.112]$, as pair 6's traced contribution falls from 0.059 to 0.010. The wrong-expert control is a consistency check on this account: with the routing frozen, the HarmBench prompts still reach the unedited pair 5, and training pair 5's records into the privacy expert moves HarmBench by $-0.005$ $[-0.016,0.000]$ and StrongREJECT by $+0.002$ $[0.000,+0.005]$. A jointly trained dense gate offers no such handle, since no single component of it serves these prompts; on the same corpus it scores 0.429 on HarmBench, $0.414$ $[0.340,0.487]$ above the edited game.

\paragraph{Reset, more experts and shorter edits.}
Table~\ref{tab:expert-edit-followup} repeats the trace-then-edit cycle on further arms of the same split, with each prediction computed from the pre-edit trace before the arm was graded.
Resetting pair 5 to the base model, with no retraining at all, gives 0.063 on HarmBench and 0.128 on StrongREJECT against predictions of 0.068 and 0.139, and an IFEval of 0.314 against a predicted 0.315.
The reset is worse than the one-epoch edit by 0.047 $[0.021,0.079]$ on HarmBench and 0.029 $[0.002,0.065]$ on StrongREJECT, so the edit does more than remove what pair 5 had learned.
The trace leaves 0.059 of the StrongREJECT score with pair 6 and 0.037 with pair 4.
Editing pairs 5 and 6 lowers StrongREJECT to 0.050, against a prediction of about 0.04 and $-0.049$ $[-0.100,-0.008]$ from the pair-5 edit, and adding pair 4 reaches the scores of the fully retrained game on both suites.
The edit also needs little training.
A quarter epoch already gives 0.016 on HarmBench and 0.107 on StrongREJECT ($+0.008$ $[-0.007,+0.032]$ against one epoch), half an epoch gives 0.016 and 0.096, and the re-traced StrongREJECT contribution of pair 5 falls from 0.359 to 0.011, 0.001 and 0.004 at a quarter, a half and one epoch.
The IFEval scores of the quarter-epoch, two-expert and three-expert edits are within two points of the original, the size of the one-epoch edit's interval, so we see no instruction-following cost.
The XSTest proxy shows the price of editing more experts: refusal of safe prompts rises from 0.068 with one edited expert to 0.309 with two, 0.326 with three and 0.369 with all seven retrained.

\begin{table}[h]
  \centering
  \caption{Further edits of the seed-0 original-corpus game with the router frozen, all graded on one host, one training run per row. Recipe as in Table~\ref{tab:expert-edit}. Lower is better for HarmBench, StrongREJECT and both XSTest columns, higher for IFEval. XSTest is a lexical refusal proxy and gives direction and rough size only: ``safe refused'' is refusal of safe prompts and ``unsafe complied'' is compliance with unsafe ones.}
  \label{tab:expert-edit-followup}
  \small
  \setlength{\tabcolsep}{4pt}
  \resizebox{\linewidth}{!}{\begin{tabular}{@{}lccccc@{}}
    \toprule
    & & & & \multicolumn{2}{c}{XSTest} \\
    \cmidrule(l){5-6}
    Model & HarmBench & StrongREJECT & IFEval & Safe refused & Unsafe complied \\
    \midrule
    \method{}, original corpus & 0.508 & 0.455 & 0.309 & 0.004 & 0.950 \\
    Pair 5 reset to the base model & 0.063 & 0.128 & 0.314 & 0.106 & 0.829 \\
    Pair 5 edited, quarter epoch & 0.016 & 0.107 & 0.290 & 0.106 & 0.819 \\
    \oursrow Pair 5 edited, one epoch & 0.016 & 0.099 & 0.298 & 0.068 & 0.819 \\
    Pairs 5 and 6 edited & 0.016 & 0.050 & 0.299 & 0.309 & 0.121 \\
    Pairs 4, 5 and 6 edited & 0.000 & 0.016 & 0.301 & 0.326 & 0.106 \\
    \midrule
    \method{}, every expert retrained on safe responses & 0.000 & 0.017 & 0.253 & 0.369 & 0.055 \\
    \bottomrule
  \end{tabular}}
\end{table}

\paragraph{Retraining the whole corpus on the same targets.}
The alternative to an edit is to retrain every model on a corpus whose harm sources carry safe responses, with the rest of the corpus and the unmasked loss unchanged (Table~\ref{tab:harm-only-retrain}).
All three retrained models reach 0.01 to 0.02 on HarmBench.
Against the original model graded on the same host, the game changes by $-0.487$ $[-0.560,-0.414]$ on HarmBench and $-0.421$ (standard error 0.050) on StrongREJECT, the dense gate by $-0.408$ (0.037) and $-0.476$ (0.048), and the pooled adapter by $-0.613$ (0.035) and $-0.553$ (0.047).
Instruction following is kept by the game ($-0.017$ $[-0.052,+0.018]$) and by the dense gate ($-0.028$ $[-0.067,+0.009]$), and the routed in-distribution gain over the pooled adapter is unchanged ($+0.125$ against $+0.117$ for the game, $+0.139$ against $+0.137$ for the dense gate).
On HarmBench the retrained game is within $-0.010$ $[-0.042,+0.021]$ of the one-expert edit that uses the same harm-source targets.
The cost is over-refusal: every retrained model refuses a third or more of the safe XSTest prompts, against 0.047 for that one-expert edit.

\begin{table}[h]
  \centering
  \caption{Whole-corpus retraining with safe responses substituted for the three harm sources only, seed-0 split, one training run per row. Columns as in Table~\ref{tab:expert-edit-followup}. Cells marked $^{\dagger}$ were graded on a second host; each original-corpus value is the one graded on the same host as the retrained cell above it, so some differ from Table~\ref{tab:expert-edit} by up to 0.03.}
  \label{tab:harm-only-retrain}
  \small
  \setlength{\tabcolsep}{4pt}
  \resizebox{\linewidth}{!}{\begin{tabular}{@{}lccccc@{}}
    \toprule
    & & & & \multicolumn{2}{c}{XSTest} \\
    \cmidrule(l){5-6}
    Model & HarmBench & StrongREJECT & IFEval & Safe refused & Unsafe complied \\
    \midrule
    \oursrow \method{}, harm sources retrained & 0.021 & 0.035$^{\dagger}$ & 0.292$^{\dagger}$ & 0.335 & 0.060 \\
    \method{}, original corpus & 0.508 & 0.456$^{\dagger}$ & 0.309$^{\dagger}$ & 0.004 & 0.950 \\
    \midrule
    Dense gate, harm sources retrained & 0.011$^{\dagger}$ & 0.014$^{\dagger}$ & 0.266 & 0.348 & 0.080 \\
    Dense gate, original corpus & 0.419$^{\dagger}$ & 0.489$^{\dagger}$ & 0.294 & 0.000 & 0.970 \\
    \midrule
    Pooled adapter, harm sources retrained & 0.016$^{\dagger}$ & 0.037$^{\dagger}$ & 0.261 & 0.411 & 0.055 \\
    Pooled adapter, original corpus & 0.628$^{\dagger}$ & 0.590$^{\dagger}$ & 0.214 & 0.004 & 0.980 \\
    \bottomrule
  \end{tabular}}
\end{table}

\paragraph{A blind replicate on a second split.}
We repeated the study on a second data split with the predictions written down before any edit was trained (Table~\ref{tab:expert-edit-replicate}).
No safe-response corpus exists for this split, so the edits use the harm-source recipe.
The unedited game scores 0.598 on HarmBench and 0.491 on StrongREJECT.
Editing pair 5 for one epoch lowers them to 0.046 and 0.138, and editing pairs 4, 5 and 6 to 0.046 and 0.039.
The wrong-expert control does not move, as predicted.
The edits fall short of their predictions, which were 0.010 and 0.112 for the one-epoch edit and 0.000 on both suites for the three-expert edit, and every edit arm misses the tolerance of 0.02 set in advance: the edited experts keep 5 to 8\% of their harm.
Three epochs do not close the gap ($-0.005$ $[-0.031,+0.021]$ and $-0.020$ $[-0.061,+0.010]$ against one epoch).
On the XSTest proxy the three-expert edit raises refusal of safe prompts from 0.008 to 0.388 and lowers compliance with unsafe prompts from 0.965 to 0.090.
The edit therefore replicates in direction and size on a second split and the control stays unmoved, while the trace overstates how completely an edited expert is cleaned.

\begin{table}[h]
  \centering
  \caption{Blind replicate of the expert edit on a second data split, harm-source recipe, all graded on one host, one training run per row (lower is better). Changes are paired differences from the unedited game of that split (HarmBench 0.598, StrongREJECT 0.491) with bootstrap 95\% intervals.}
  \label{tab:expert-edit-replicate}
  \small
  \setlength{\tabcolsep}{4pt}
  \resizebox{\linewidth}{!}{\begin{tabular}{@{}lcccc@{}}
    \toprule
    Model & HarmBench & Change & StrongREJECT & Change \\
    \midrule
    \oursrow Pair 5 edited, one epoch & 0.046 & $-0.552$ $[-0.624,-0.474]$ & 0.138 & $-0.354$ $[-0.449,-0.255]$ \\
    Pair 5 edited, three epochs & 0.041 & $-0.557$ $[-0.629,-0.485]$ & 0.118 & $-0.373$ $[-0.465,-0.282]$ \\
    Pairs 4, 5 and 6 edited & 0.046 & $-0.552$ $[-0.624,-0.474]$ & 0.039 & $-0.453$ $[-0.541,-0.361]$ \\
    Wrong-expert control (pair 0 on pair 5's records) & 0.608 & $+0.010$ $[-0.031,+0.052]$ & 0.521 & $+0.030$ $[-0.006,+0.075]$ \\
    \midrule
    Pooled adapter, unedited & 0.582 & $-0.015$ $[-0.093,+0.062]$ & 0.519 & $+0.028$ $[-0.055,+0.110]$ \\
    Dense gate, unedited & 0.294 & $-0.304$ $[-0.387,-0.222]$ & 0.394 & $-0.097$ $[-0.181,-0.014]$ \\
    \bottomrule
  \end{tabular}}
\end{table}

\section{Out-of-distribution evaluation}
\label{app:ood}

The trained models are scored on data that no arm saw, with every prompt routed through the unchanged trait space. Unless stated otherwise the models are the seven-pair $T=0.35$ game, a dense softmax gate and the pooled adapter of the seed-0 split.

\paragraph{Far from the training distribution.}
None of the seven training benchmarks is mathematical. We score all 1,319 GSM8K test problems \citep{cobbe2021gsm8k} two ways: the NLL of the reference solution with the question as context, and answer accuracy from greedy decoding of 512 new tokens with an exact match on the final number. The three training corpora are the original one and the two corrected ones of Appendix~\ref{app:safety}. Table~\ref{tab:ood-gsm8k} shows that fine-tuning on this corpus costs every model most of the base model's ability, and that the game keeps the most. On the original corpus its mixture has the lowest solution NLL of the fine-tuned models, and its routed expert answers more than twice as many problems as the pooled adapter. With safe responses the pooled adapter has the lowest solution NLL while the game stays the most accurate. The game's hindsight-best expert, chosen per problem with the reference solution, has a solution NLL of 0.669 on the original corpus, 0.676 with unsafe targets removed and 0.661 with safe responses. It is an upper bound on routing with no decoding counterpart, and it nearly matches the base model, so some experts keep their mathematical ability, and the router, which has never seen such prompts, sends 55\% of the problems to one expert that is good but not the best.

\begin{table}[h]
  \centering
  \caption{GSM8K, 1,319 test problems, seed-0 split. Solution NLL in nats per token (lower is better) and answer accuracy (higher is better) for models trained on the original corpus, with unsafe targets removed, and with safe responses and a prompt-masked loss. The base model scores 0.668 and 0.706.}
  \label{tab:ood-gsm8k}
  \small
  \begin{tabular}{@{}lcccccc@{}}
    \toprule
    & \multicolumn{3}{c}{Solution NLL} & \multicolumn{3}{c}{Answer accuracy} \\
    \cmidrule(lr){2-4}\cmidrule(lr){5-7}
    Model & Original & Removed & Safe resp. & Original & Removed & Safe resp. \\
    \midrule
    \oursrow \method{}, token-level mixture & 0.716 & 0.735 & 0.775 & 0.425 & 0.413 & 0.403 \\
    \oursrow \method{}, one routed expert & 0.723 & 0.749 & 0.787 & 0.428 & 0.419 & 0.397 \\
    Pooled adapter & 0.742 & 0.786 & 0.754 & 0.194 & 0.192 & 0.350 \\
    Dense softmax gate & 0.792 & 0.755 & 0.773 & 0.177 & 0.278 & 0.300 \\
    \bottomrule
  \end{tabular}
\end{table}

The accuracy result holds over three data splits of the original corpus. With the train-only trait space of Appendix~\ref{app:inductive} on every split and the dense gate at a $1\times16$ micro-batch, the game's mixture answers $0.414\pm0.056$ of the problems against $0.207\pm0.031$ for the pooled adapter and $0.222\pm0.020$ for the dense gate, a paired lead of $0.207\pm0.033$ over the pooled adapter and $0.192\pm0.044$ over the dense gate. The generations explain the gap. The pooled adapter and the dense gate answer in one short line (median 20 and 23 new tokens against 283 for the base model), whereas the game's experts keep more of the worked solution (median 70--73 tokens). The game also inherits one failure from the corpus: on every split the pair trained on the privacy data receives 118 to 176 name-heavy problems and rewrites them with the entities masked instead of solving them. On pseudo-Fisher axes the game's mixture degrades least on solution NLL on all three splits, better than the pooled adapter by $0.043\pm0.013$ nats and than its dense gate by $0.032\pm0.014$, and on the seed-0 split it answers 0.305 of the problems against 0.215 for its dense gate.

\paragraph{New datasets of the trained kinds.}
For each training category we take a public dataset that was never used in training and remove prompts that duplicate a training prompt exactly: PKU-SafeRLHF \citep{ji2024pkusaferlhf}, TruthfulQA, a jailbreak-classification set (\texttt{jackhhao/jailbreak-classification}), PII-Masking-300k from the publisher of the privacy training data, XSTest, \texttt{deepset/prompt-injections} and CoCoNot \citep{brahman2024coconot}. Table~\ref{tab:ood-near} scores them with the full-chat-text NLL used throughout. Both mixtures beat the pooled adapter on average, the game's by 0.045 nats and the dense gate's by 0.062, so the dense gate is ahead of the game by 0.018. Routing does not follow a dataset's nominal category. Except for the privacy set, only 3--26\% of a new dataset's prompts reach the expert that serves its category in distribution, and on XSTest, the injection set and CoCoNot the experts the router chooses instead are better than the category's own expert, so the new prompts resemble other regions of the trait space. A nearest-centroid check agrees: test prompts land with their own benchmark 92\% of the time on labelled axes, and new datasets land with their nominal category 26\% of the time, with 25--29\% on the label-free spaces of Appendix~\ref{app:label-free}. Each category is learned from one source, and a new dataset of the same kind is often closer to a different source.

\begin{table}[h]
  \centering
  \caption{Seven unseen datasets of the trained kinds, original-corpus models of the seed-0 split, full-chat-text NLL (lower is better). The last column is the share of prompts routed to the expert that serves the dataset's category, with the in-distribution share in parentheses.}
  \label{tab:ood-near}
  \small
  \setlength{\tabcolsep}{4pt}
  \resizebox{\linewidth}{!}{\begin{tabular}{@{}llrcccc@{}}
    \toprule
    New dataset & Trained category & $n$ & Pooled & \method{} mixture & Dense gate & To category expert \\
    \midrule
    PKU-SafeRLHF & harm & 955 & 1.901 & 1.931 & 1.886 & 0.26 (0.84) \\
    TruthfulQA & hallucination & 817 & 2.144 & 2.075 & 2.145 & 0.16 (0.98) \\
    Jailbreak classification & jailbreak & 400 & 2.662 & 2.667 & 2.666 & 0.20 (1.00) \\
    PII-Masking-300k & privacy & 1,000 & 1.607 & 1.621 & 1.634 & 0.90 (0.98) \\
    XSTest & over-refusal & 434 & 2.365 & 2.253 & 2.156 & 0.03 (0.93) \\
    deepset prompt injections & injection & 662 & 3.584 & 3.468 & 3.463 & 0.13 (0.94) \\
    CoCoNot & policy violation & 1,001 & 2.808 & 2.743 & 2.686 & 0.16 (0.69) \\
    \midrule
    Unweighted mean (total $n$) & all & 5,269 & 2.439 & 2.394 & 2.376 & 0.26 (0.91) \\
    \bottomrule
  \end{tabular}}
\end{table}

The full-text metric includes the prompt, which the fine-tuned models learn to model. On the response tokens of the three datasets that have reference responses (PKU-SafeRLHF, TruthfulQA and PII-Masking-300k, 2,772 records), no mixture beats the pooled adapter on average on the original corpus: the pooled adapter scores 1.331, the dense gate 1.334 and the game's mixture 1.342, against 1.543 for the base model. The game's hindsight-best expert scores 1.219, so the experts carry variety on these responses that neither router reaches on new data. On pseudo-Fisher axes over three data splits, both mixtures beat the pooled adapter on full text and the dense gate leads the game by $0.018\pm0.012$ nats, the labelled-axes picture again.

\paragraph{Why the game loses on new harm data.}
PKU-SafeRLHF is the one new dataset on which the game trails the pooled adapter in Table~\ref{tab:ood-near}, and the per-record scores locate the loss in the routing. On PKU response tokens the harm expert alone matches the pooled adapter (1.8916 against 1.8920) and is the best expert for 72\% of the records, but only 26\% of them are routed to it, against 84\% of the BeaverTails test prompts. The game's mixture therefore scores 1.9421, while its hindsight-best expert scores 1.8820. The label-free pseudo-Fisher axes route 50\% of PKU to the harm expert and lose less (1.9219). Two changes to the labelled trait space, each with the game and its pooled adapter retrained, do not move the PKU prompts (Table~\ref{tab:ood-harm}). The first rebuilds the harm category from two sources, 2,500 BeaverTails and 2,500 Aegis~2.0 records \citep{ghosh2025aegis2}. The two-source harm expert now beats its own pooled adapter on PKU responses, but the routed share does not rise and the mixture still trails. The second fits every labelled axis on prompt embeddings, where the harm and hallucination axes are otherwise fitted on responses. It keeps the in-distribution gain and again leaves the routed share unchanged or lower. In every variant the PKU prompts sit at a mean harm coordinate of 0.21--0.22, against 0.33 for the BeaverTails test prompts. The failure on a new harm source therefore lies in where its prompts sit in the encoder's embedding, and neither the text the axes are fitted on nor the number of sources per category changes it.

\begin{table}[h]
  \centering
  \caption{PKU-SafeRLHF under changes to the labelled trait space, seed-0 split, train-only axes, response-token NLL (lower is better) on the 942 PKU prompts absent from every compared corpus. Routed is the share of PKU prompts whose largest share goes to the harm expert. The last column is the game's expected gain over its pooled adapter on the in-distribution test pool, full chat text.}
  \label{tab:ood-harm}
  \small
  \setlength{\tabcolsep}{4pt}
  \resizebox{\linewidth}{!}{\begin{tabular}{@{}llcccccc@{}}
    \toprule
    Harm sources & Labelled axes fitted on & Pooled & \method{} mixture & Harm expert & Oracle & Routed & In-dist.\ gain \\
    \midrule
    One & responses and prompts & 1.8969 & 1.9511 & 1.8972 & 1.8862 & 0.23 & 0.105 \\
    One & prompts only & 1.8969 & 1.9543 & 1.8954 & 1.8830 & 0.21 & 0.111 \\
    Two & responses and prompts & 1.9248 & 1.9591 & 1.9075 & 1.8964 & 0.24 & 0.104 \\
    Two & prompts only & 1.9248 & 1.9547 & 1.9069 & 1.8954 & 0.18 & 0.106 \\
    \bottomrule
  \end{tabular}}
\end{table}

\section{Reproduction}
\label{app:repro}

\paragraph{Environment and entry point.}
The implementation is a Python package requiring Python~3.10 or later; dependency constraints and optional training, plotting, testing, and behavioral-evaluation extras are specified in \texttt{pyproject.toml}. All experiments use the same YAML-driven pipeline rather than experiment-specific scripts. The principal run can be validated and executed with:
\begin{verbatim}
python -m pip install -e \
  ".[ml,configs,figures,test,behavioral]"
python -m pytest
python -m infl_ens.pipeline \
  --config configs/experiments/seven_axis_theory_eq.yaml \
  --dry-run
python -m infl_ens.pipeline \
  --config configs/experiments/seven_axis_theory_eq.yaml
\end{verbatim}
The dry run resolves every layered arm configuration and reports its output directory and trait-space fingerprint without loading a model. Experiment files for the parameter sweeps, model ablations, baselines, and behavioral evaluation are colocated under \texttt{configs/experiments/}; each declares its arms, stage order, evaluation partition, and figure builders. Configurations that define a smoke block can additionally be checked with \texttt{--smoke} before a full GPU run.

\paragraph{Data identity, seeds, and cached geometry.}
The shared split manifest stores stable record identifiers and the within-benchmark train/validation/test assignment, so controlled arms consume identical examples rather than merely matching split sizes. Training, splitting, routing, and generation seeds are fields of the resolved configuration. Each completed run writes the fully flattened \texttt{resolved\_config.yaml}; this file, rather than the compositional arm fragment, is the authoritative record of model, LoRA, game, and evaluation settings. The principal box-coordinate trait space uses Qwen3-Embedding-8B-AWQ-INT4 and has configuration fingerprint \texttt{3b42c68a8dd334c5}; all corresponding arms validate this identity before loading the shared cache.

\paragraph{Artifact contract.}
Every training run writes \texttt{history.json} with its round states and \texttt{resolved\_config.yaml} with its effective configuration; runs using the shared manifest also write \texttt{data\_split.json}. Final routing summaries are reported in round-11 \texttt{routing\_ensemble\_diagnostics.json}. The underlying \texttt{merge\_nll.npy}, \texttt{g\_merge.npy}, and \texttt{token\_counts.npy} arrays are sufficient to recompute the reported commitment, expected-NLL, and probability-mixture estimators without reloading model checkpoints. Behavioral generation and score artifacts are schema-versioned and record ordered-input, checkpoint, trait-coordinate, and content digests; the reported behavioral tables use these artifacts rather than reparsing model logs. Experiment-level \texttt{stage\_status.json} records stage completion and timing, while figure generation reads the saved artifacts without rerunning training.

\paragraph{Numerical reproducibility caveat.}
The four-bit AWQ trait encoder is not bitwise invariant across GPU architectures. In a 206-prompt cross-hardware check spanning L40S, A5500, and H200 devices, recomputing coordinates changed the hard-routing argmax for 2.4--3.4\% of prompts. Exact reproduction therefore requires reuse of the original trait-space and behavioral-coordinate caches, not merely the same encoder model identifier. Behavioral coordinate caches are keyed by both the suite digest and trait-space fingerprint, and their content hash is included in generation provenance; rebuilding coordinates on different hardware should be treated as a robustness replication rather than a bitwise reproduction.
 
\end{document}